\documentclass{article}

\usepackage[preprint]{neurips_2023}

\usepackage[utf8]{inputenc} 
\usepackage[T1]{fontenc}    
\usepackage{amsfonts,amsmath,amssymb,amsthm}
\usepackage{physics,mathtools,bm,mathrsfs,mleftright}
\usepackage{xcolor,enumerate,autobreak}
\usepackage{wrapfig,subfigure,multirow}
\usepackage{url}            
\usepackage{booktabs}       
\usepackage{nicefrac}       
\usepackage{microtype}      
\usepackage{hyperref,prettyref}

\theoremstyle{plain}
\newtheorem{theorem}{Theorem}[section]
\newtheorem{lemma}[theorem]{Lemma}

\theoremstyle{definition}
\newtheorem{proposition}[theorem]{Proposition}
\newtheorem{definition}[theorem]{Definition}
\newtheorem{remark}[theorem]{Remark}

\newtheorem{assumption}{Assumption}

\newrefformat{eq}{\textup{(\ref{#1})}}
\newrefformat{lem}{Lemma~\textup{\ref{#1}}}
\newrefformat{thm}{Theorem~\textup{\ref{#1}}}
\newrefformat{cor}{Corollary~\textup{\ref{#1}}}
\newrefformat{prop}{Proposition~\textup{\ref{#1}}}
\newrefformat{assu}{Assumption~\textup{\ref{#1}}}
\newrefformat{rem}{Remark~\textup{\ref{#1}}}
\newrefformat{def}{Definition~\textup{\ref{#1}}}
\newrefformat{sec}{Section~\textup{\ref{#1}}}
\newrefformat{subsec}{Section~\textup{\ref{#1}}}

\newcommand{\ep}{\varepsilon}
\newcommand{\R}{\mathbb{R}}
\newcommand{\E}{\mathbb{E}}
\newcommand{\K}{\mathbb{K}}

\newcommand{\caF}{\mathcal{F}}

\newcommand{\caH}{\mathcal{H}}

\newcommand{\caX}{\mathcal{X}}

\newcommand{\bbP}{\mathbb{P}}

\DeclareMathOperator*{\argmin}{arg\,min}

\DeclareMathOperator{\ran}{Ran}

\newcommand{\poly}{\mathrm{poly}}
\newcommand{\ang}[1]{\left\langle{#1}\right\rangle}

\providecommand{\cref}{\prettyref}

\usepackage{bbding}
\usepackage{verbatim}
\usepackage{float}

\title{On the Asymptotic Learning Curves of Kernel Ridge Regression under Power-law Decay}

\author{Yicheng Li,~ Haobo Zhang \\
Center for Statistical Science, Department of Industrial Engineering \\ Tsinghua University, Beijing, China \\
\texttt{\{liyc22,zhang-hb21\}@mails.tsinghua.edu.cn} \\
\And
Qian Lin \thanks{Qian Lin also affiliates with Beijing Academy of Artificial Intelligence, Beijing, China} \\
Center for Statistical Science, Department of Industrial Engineering \\ Tsinghua University, Beijing, China \\
\texttt{qianlin@tsinghua.edu.cn}
}

\begin{document}

  \maketitle

  \begin{abstract}
    The widely observed `benign overfitting phenomenon' in the neural network literature raises the challenge to the `bias-variance trade-off' doctrine in the statistical learning theory.
Since the generalization ability of the `lazy trained' over-parametrized neural network can be well approximated by that of the neural tangent kernel regression,
the curve of the excess risk (namely, the learning curve) of kernel ridge regression attracts increasing attention recently.
However, most recent arguments on the learning curve are heuristic and are based on the `Gaussian design' assumption.
In this paper, under mild and more realistic assumptions, we rigorously provide a full characterization of the learning curve:
elaborating the effect and the interplay of the choice of the regularization parameter, the source condition and the noise.
In particular, our results suggest that the `benign overfitting phenomenon' exists in very wide neural networks only when the noise level is small.

  \end{abstract}

  \section{Introduction}\label{sec:introduction}
  Kernel methods, in particular kernel ridge regression (KRR), have been one of the most popular algorithms in machine learning.
Its optimality under various settings has been an active topic since \citet{caponnetto2007_OptimalRates,andreaschristmann2008_SupportVector}.
The renaissance of kernel methods arising from the neural tangent kernel (NTK) theory
~\citep{jacot2018_NeuralTangent}, which shows that over-parametrized neural networks can be well approximated by certain kernel regression with the corresponding NTK,
has posed further challenges about the interplay of generalization, regularization and noise level.
For example, it has been observed empirically that over-parametrized neural networks can fit any data perfectly but also generalize well
~\citep{zhang2017_UnderstandingDeep}, 
which contradicts to our traditional belief of bias-variance trade-off~\citep{vapnik1999_NatureStatistical}.

The aforementioned `benign overfitting phenomenon' that overfitted neural networks generalize well attracts lots of attention recently.
Researchers provide various explanations to reconcile the contradiction between it and the bias-variance trade-off principle.
For example, \citet{belkin2019_ReconcilingModern} proposed the `double descent theory' to explain why large model can generalize well;
some other works (e.g., \citet{liang2020_JustInterpolate})
argued that kernel interpolating estimators can generalize well in high dimensional settings.
In contrast to the `benign overfitting phenomenon', several other works (e.g., \citet{rakhlin2018_ConsistencyInterpolation,li2023_KernelInterpolation})
recently showed that kernel interpolation can not generalize in traditional fixed dimension setting.
In order to understand the `benign overfitting phenomenon',
it would be of great interest to characterize the learning curve:
the curve of the exact order of the generalization error of a certain algorithm (e.g., KRR) varying with respect to different choices of regularization parameters.

Recently, several works (e.g., \citet{bordelon2020_SpectrumDependent,cui2021_GeneralizationError})
depicted the learning curve of KRR under the Gaussian design assumption that the eigenfunctions (see \cref{eq:Mercer}) are i.i.d.\ Gaussian random functions.
Though it is easy to figure out that the Gaussian design assumption can not be true in most scenarios,
with some heuristic arguments, \citet{cui2021_GeneralizationError}  provide a description of the learning curves of KRR
with respect to the regularization, source condition and noise levels.
These works offered us some insights on the learning curve of KRR which strongly suggests that the learning curve should be U-shaped if the observations are noisy or monotone decreasing if the observations are noiseless.

In this paper, we consider the learning curves of KRR under the usual settings (without the Gaussian design assumption).
Under mild assumptions, we rigorously prove the asymptotic rates of the excess risk, including both upper and lower bounds.
These rates show the interplay of the eigenvalue decay of the kernel, the relative smoothness of the regression function,
the noise and the choice of the regularization parameter.
As a result, we obtain the traditional U-shaped learning curve for the noisy observation case and
a monotone decreasing learning curve for the noiseless case,
providing a full picture of the generalization of KRR in the asymptotic sense.
Combined with the NTK theory, our results may also suggest that `the benign overfitting phenomenon' may not exist if one trains a very wide neural network.


\subsection{Our contributions}

The main contribution of this paper is that
we remove the unrealistic Gaussian design assumption in previous non-rigorous works~\citep{bordelon2020_SpectrumDependent,cui2021_GeneralizationError}
and provide mathematically solid proof of the exact asymptotic rates of KRR with matching upper and lower bounds.

To be precise, let us introduce the quantities $\lambda$, the regularization parameter in \cref{eq:KRR_Def};
$\beta$, the eigenvalue decay rate in \cref{eq:EDR}, which characterizes the span of the underlying reproducing kernel Hilbert space (RKHS);
and $s$, the smoothness index in \cref{eq:fHs}, describes the relative smoothness of the regression function with respect to the RKHS\@.
Here we note that larger $\beta$ implies better regularity the RKHS and also larger $s$ also implies better relative smoothness.
Then, the asymptotic rates of the generalization error (excess risk) $R(\lambda)$ in the noisy case is roughly
\begin{align*}
  R(\lambda) =
  \begin{cases}
    \Theta\mleft( \lambda^{\min(s,2)} +  \sigma^2\lambda^{-1/\beta} / n \mright), & \qif \lambda = \Omega(n^{-\beta}); \\
    \Omega(\sigma^2), & \qif \lambda = O(n^{-\beta});
  \end{cases}
\end{align*}
where $n$ is the number of the samples and $\sigma^2$ is the noise level.
This result justifies the traditional U-shaped learning curve (see also \cref{fig:LearningCurve}) with respect to the regularization parameter.

For the technical part,
we use the bias-variance decomposition and determine the exact rates of the both terms.
Since the variance term was already considered in \citet{li2023_KernelInterpolation}, the main focus of this work is the bias term.
Our technical contributions include:
\begin{itemize}
  \item When the regularization parameter $\lambda$ is not so small, that is, $\lambda = \Omega(n^{-\beta})$,
  we provide sharp estimates of the asymptotic orders (\cref{lem:BiasAsymp}) of the bias term with both upper and lower bounds.
  Our result holds for both the well-specified case ($s \geq 1$) and the mis-specified case ($s \in (0,1)$),
  which improves the upper bounds given in \citet{zhang2023_OptimalityMisspecified}.
  \item We further show an upper bound (\cref{lem:BiasUpperInterpolating}) of the bias term in the nearly interpolating case,
  i.e.,  $\lambda = O(n^{-\beta})$.
  The upper bound is tight and matches the information-theoretic lower bound provided in \cref{prop:BiasLower}.
  \item Combining these results, we provide learning curves of KRR for both the noisy case (\cref{thm:LearningCurve_Noisy})
  and the noiseless case (\cref{thm:LearningCurve_Noiseless}).
  The results justify our traditional belief of the bias-variance trade-off principle.
  \item Our new techniques can also be generalized to other settings and might be of independent interest.
\end{itemize}

\subsection{Related works}\label{subsec:related}
The optimality of kernel ridge regression has been studied extensively~\citep{caponnetto2007_OptimalRates,steinwart2009_OptimalRates,fischer2020_SobolevNorm,zhang2023_OptimalityMisspecified}.
\citet{caponnetto2007_OptimalRates} provided the classical optimality result of KRR in the well-specified case
and the subsequent works further considered the mis-specified case.
However, these works only provided an upper bound and the worst-case (minimax) lower bound,
which are not sufficient for determining the precise learning curve.
In order to answer the ``benign overfitting'' phenomenon~\citep{bartlett2020_BenignOverfitting,liang2020_JustInterpolate},
several works~\citep{rakhlin2018_ConsistencyInterpolation,buchholz2022_KernelInterpolation,beaglehole2022_KernelRidgeless}
tried to provide a lower bound for the kernel interpolation, which is a limiting case of KRR\@,
but these works only focused on particular kernels and their techniques can hardly be generalized to provide a lower bound for KRR\@.

Another line of recent works considered the generalization performance of KRR under the Gaussian design assumption of the eigenfunctions
~\citep{bordelon2020_SpectrumDependent,jacot2020_KernelAlignment,cui2021_GeneralizationError,mallinar2022_BenignTempered}.
In particular, the learning curves of KRR was described in \citet{bordelon2020_SpectrumDependent,cui2021_GeneralizationError},
but heuristic arguments are also made in addition to the unrealistic Gaussian design assumption.
Though the heuristic arguments are inspirational, a rigorous proof is indispensable if one plans to perform further investigations.
In this work, we provide the first rigorous proof for most scenarios of the smoothness $s$,
eigenvalue decay rate $\beta$, noise level $\sigma^2$ and the regularization parameter $\lambda$ based on the most common/realistic assumptions.


Recently, in order to show the so-called ``saturation effect'' in KRR,
\citet{li2023_SaturationEffect} proved the exact asymptotic order of both the bias and the variance term
when the regression function is very smooth and the regularization parameter $\lambda$ is relatively large.
Inspired by their analysis, \citet{li2023_KernelInterpolation} showed the exact orders of the variance term.
Our work further determines the orders of the bias term, completing the full learning curve or KRR\@.


KRR is also connected with Gaussian process regression~\citep{kanagawa2018_GaussianProcesses}.
\citet{jin2021_LearningCurves} claimed to establish the learning curves for Gaussian process regression and thus for KRR\@.
However, as pointed out in \citet{zhang2023_OptimalityMisspecifieda}, there is a gap in their argument.
Moreover, their results are also more restrictive than ours, see \cref{subsec:Discussion} for a comparison.

\paragraph{Notations}
We write $L^p(\caX,\dd \mu)$ for the Lebesgue space and sometimes abbreviate it as $L^p$.
We use asymptotic notations $O(\cdot),~o(\cdot),~\Omega(\cdot)$ and $\Theta(\cdot)$, and use $\tilde{\Theta}(\cdot)$ to suppress logarithm terms.
We also write $a_n \asymp b_n$ for $a_n = \Theta(b_n)$.
We will also use the probability versions of the asymptotic notations such as $O_{\bbP}(\cdot)$.
Moreover, to present the results more clearly, we denote $a_n = O^{\poly}(b_n)$ if $a_n = O(n^p b_n)$ for any $p > 0$,
$a_n = \Omega^{\poly}(b_n)$ if $a_n = \Omega(n^{-p} b_n)$ for any $p > 0$, $a_n = \Theta^{\poly}(b_n)$ if $a_n = O^{\poly}(b_n)$, and $a_n = \Omega^{\poly}(b_n)$;
and we add a subscript $ _\bbP$ for their probability versions.

  \section{Preliminaries}\label{sec:preliminaries}
  
Let $\caX\subset \R^{d}$ be compact and $\rho$ be a probability measure on $\caX\times \R$,
whose marginal distribution on $\caX$ is denoted by $\mu$.
Suppose that we are given $n$ i.i.d.\ samples $(x_1,y_1),\dots,(x_n,y_n)$ from $\rho$.
Let $k$ be a continuous positive definite kernel $k$ over $\caX$ and $\caH$ be the separable reproducing kernel Hilbert space (RKHS) associated with $k$.
Then, kernel ridge regression (KRR) obtains the regressor $\hat{f}_\lambda$ via the following convex optimization problem
\begin{align}
  \label{eq:KRR_Def}
    \hat{f}_{\lambda} &= \argmin_{f \in \caH} \left( \frac{1}{n}\sum_{i=1}^n (y_i - f(x_i))^2 + \lambda \norm{f}_{\caH}^2 \right),
\end{align}
where $\lambda > 0$ is the regularization parameter.
Let us denote $X = (x_1,\dots,x_n)$ and $\bm{y} = (y_1,\dots,y_n)^T$.
A closed form of \cref{eq:KRR_Def} can be provided by the representer theorem~\citep{andreaschristmann2008_SupportVector}:
\begin{align}
  \label{eq:KRR_Matrix}
  \hat{f}_{\lambda}(x)&= \K(x,X)(\K(X,X)+n \lambda)^{-1}\bm{y}
\end{align}
where $\K(x,X) = \left( k(x,x_1),\dots,k(x,x_n) \right)$ and $\K(X,X) = \big( k(x_i,x_j) \big)_{n\times n}$.

In terms of the generalization performance of $\hat{f}_{\lambda}$, we consider the excess risk with respect to the squared loss
\begin{align}
  \label{eq:ExcessRisk}
    \E_{x \sim \mu}\left[ \hat{f}_\lambda(x) - f^*_\rho(x) \right]^2 = \norm{\hat{f}_\lambda - f^*_\rho}_{L^2(\caX,\dd \mu)}^2,
\end{align}
where $f^*_\rho(x) \coloneqq \E_{\rho}[y\mid x]$ is the conditional expectation and is also referred to as the regression function.
We aim to provide asymptotic orders of \cref{eq:ExcessRisk} with respect to $n$.

\subsection{The integral operator}
We will introduce the integral operator, which is crucial for the analysis, as the previous works~\citep{caponnetto2007_OptimalRates,lin2018_OptimalRates}.
Denote by $\mu$ the marginal probability measure of $\rho$ on $\caX$.
Since $k$ is continuous and $\caX$ is compact, let us assume $\sup_{x \in \caX}k(x,x) \leq \kappa^2$.
Then, it is known~\citep{andreaschristmann2008_SupportVector,steinwart2012_MercerTheorem} that
we have the natural embedding $S_\mu : \caH \to L^2$, which is a Hilbert-Schmidt operator with Hilbert-Schmidt norm $\norm{S_\mu}_{\mathrm{HS}} \leq \kappa$.
Let $S_\mu^* : L^2 \to \caH$ be the adjoint operator of $S_\mu$ and $T = S_\mu S_\mu^* : L^2 \to L^2$.
Then, it is easy to show that $T$ is an integral operator given by
\begin{align}
(Tf)(x)
  = \int_{\mathcal{X}} k(x,y) f(y) \dd \mu(y),
\end{align}
and it is self-adjoint, positive and trace-class (thus compact) with trace norm $\norm{T}_1 \leq \kappa^2$~\citep{caponnetto2007_OptimalRates,steinwart2012_MercerTheorem}.
Moreover, the spectral theorem of compact self-adjoint operators and Mercer's theorem~\citep{steinwart2012_MercerTheorem} yield the decompositions
\begin{align}
  \label{eq:Mercer}
  T = \sum_{i\in N} \lambda_i \ang{\cdot,e_i}_{L^2} e_i,\quad\quad
  k(x,y) = \sum_{i\in N} \lambda_i e_i(x) e_i(y),
\end{align}
where $N \subseteq \mathbb{N}$ is an index set, $\left\{ \lambda_i \right\}_{i \in N}$ is the set of positive eigenvalues of $T$ in descending order,
and $e_i$ is the corresponding eigenfunction.
Furthermore, $\left\{ e_i \right\}_{i \in N}$ forms an orthonormal basis of $\overline{\ran S_\mu} \subseteq L^2$
and $\left\{ \lambda_i^{1/2} e_i \right\}_{i\in N}$ forms an orthonormal basis of $\overline{\ran S_\mu^*} \subseteq \caH$.

The eigenvalues $\lambda_i$ actually characterize the span of the RKHS and the interplay between $\caH$ and $\mu$.
Since we are interested in the infinite-dimensional case, we will assume $N = \mathbb{N}$ and
assume the following polynomial eigenvalue decay as in the literature
~\citep{caponnetto2007_OptimalRates,fischer2020_SobolevNorm,li2023_SaturationEffect},
which is also referred to as the capacity condition or effective dimension condition.
Larger $\beta$ implies better regularity of the functions in the RKHS\@.

\begin{assumption}[Eigenvalue decay]
  \label{assu:EDR}
  There is some $\beta > 1$ and constants $c_\beta,C_\beta > 0$ such that
  \begin{align}
    \label{eq:EDR}
    c_\beta i^{- \beta} \leq \lambda_i \leq C_\beta i^{-\beta} \quad (i = 1,2,\dots),
  \end{align}
  where $\lambda_i$ is the eigenvalue of $T$ defined in \cref{eq:Mercer}.
\end{assumption}

Such a polynomial decay is satisfied for the well-known Sobolev kernel~\citep{fischer2020_SobolevNorm}, Laplace kernel and,
of most interest, neural tangent kernels for fully-connected multilayer neural networks
~\citep{bietti2019_InductiveBias,bietti2020_DeepEquals,lai2023_GeneralizationAbility}.

\subsection{The embedding index of an RKHS}\label{subsec:examples of RKHS}
We will consider the embedding index of an RKHS to sharpen our analysis.
Let us first define the fractional power $T^s : L^2 \to L^2$ for $s \geq 0$ by
\begin{align}
  T^s(f) &= \sum_{i\in N} \lambda_i^s \ang{f,e_i}_{L^2} e_i.
\end{align}
Then, the interpolation space~\citep{steinwart2012_MercerTheorem,fischer2020_SobolevNorm,li2023_SaturationEffect} $[\caH]^s$ is define by
\begin{align}
\label{eq:InterpolationSpaceDef}
[\caH]
  ^s = \ran T^{s/2} = \left\{ \sum_{i \in N} a_i \lambda_i^{s/2} e_i ~\Big|~ \sum_{i\in N} a_i^2 < \infty \right\}
  \subseteq L^2,
\end{align}
with the norm $\norm{\sum_{i \in N} a_i \lambda_i^{s/2} e_i}_{[\caH]^s} = \left(  \sum_{i\in N} a_i^2  \right)^{1/2}$.
%
One may easily verify that $[\caH]^s$ is also a separable Hilbert space with an orthonormal basis $\left\{ \lambda_i^{s/2} e_i \right\}_{i \in N}$.
Moreover, it is clear that $[\caH]^0 = \overline{\ran S_{\mu}} \subseteq L^2$ and $[\caH]^1 = \overline{\ran S_\mu^*} \subseteq \caH$.
It can also be shown that if $s_1 > s_2 \geq 0$, the inclusions $[\caH]^{s_1} \hookrightarrow [\caH]^{s_2}$ are compact~\citep{steinwart2012_MercerTheorem}.

Now, we say $\caH$ has an embedding property of order $\alpha \in (0,1]$ if $[\mathcal{H}]^\alpha$ can be continuously embedded into $L^\infty(\caX,\dd \mu)$, that is,
the operator norm
\begin{equation}
  \label{eq:EMB}
  \norm{[\mathcal{H}]^\alpha \hookrightarrow L^{\infty}(\mathcal{X},\mu)} = M_\alpha < \infty.
\end{equation}
Moreover, \citet[Theorem 9]{fischer2020_SobolevNorm} shows that
\begin{align}
  \label{eq:EMB_And_InfNorm}
  \norm{[\mathcal{H}]^\alpha \hookrightarrow L^{\infty}(\mathcal{X},\mu)} = \norm{k^\alpha_{\mu}}_{L^\infty}
  \coloneqq \operatorname*{ess~sup}_{x \in \caX,~\mu} \sum_{i \in N} \lambda_i^{\alpha} e_i(x)^2.
\end{align}
Therefore, since $\sup_{x \in \caX}k(x,x) \leq \kappa^2$, we know that \cref{eq:EMB} always holds for $\alpha = 1$.
By the inclusion relation of interpolation spaces, it is clear that if $\caH$ has the embedding property of order $\alpha$,
then it has the embedding properties of order $\alpha'$ for any $\alpha'\geq\alpha$.
Consequently, we may introduce the following definition~\citep{zhang2023_OptimalityMisspecifieda}:
\begin{definition}
    The embedding index $\alpha_{0}$ of an RKHS $\caH$ is defined by
    \begin{align}
  \label{eq:EMB_Idx}
    \alpha_{0} = \inf\left\{ \alpha :  \norm{[\mathcal{H}]^\alpha \hookrightarrow L^{\infty}(\mathcal{X},\mu)} = M_\alpha < \infty  \right\}.
\end{align}
\end{definition}

It is shown in \citet[Lemma 10]{fischer2020_SobolevNorm} that $\alpha_0 \geq \beta$ and we assume the equality holds as the following assumption.
\begin{assumption}[Embedding index]
  \label{assu:EMB}
  The embedding index $\alpha_0 = 1/\beta$, where $\beta$ is the eigenvalue decay in \cref{eq:EDR}.
\end{assumption}

Lots of the usual RKHSs satisfy this embedding index condition.
It is shown in \citet{steinwart2009_OptimalRates} that \cref{assu:EMB} holds if the eigenfunctions are uniformly bounded,
namely $\sup_{i \in N} \norm{e_i}_{L^\infty} < \infty$.
Moreover, \cref{assu:EMB} also holds for the Sobolev RKHSs, RKHSs associated with periodic translation invariant kernels and
RKHSs associated with dot-product kernels on spheres, see \citet[Section 4]{zhang2023_OptimalityMisspecified}.

  \section{Main Results}\label{sec:main-results}
  \begin{figure}[htb]
  \centering
  \subfigure{
    \begin{minipage}[t]{0.5\linewidth}
      \centering
      \includegraphics[width=1\linewidth]{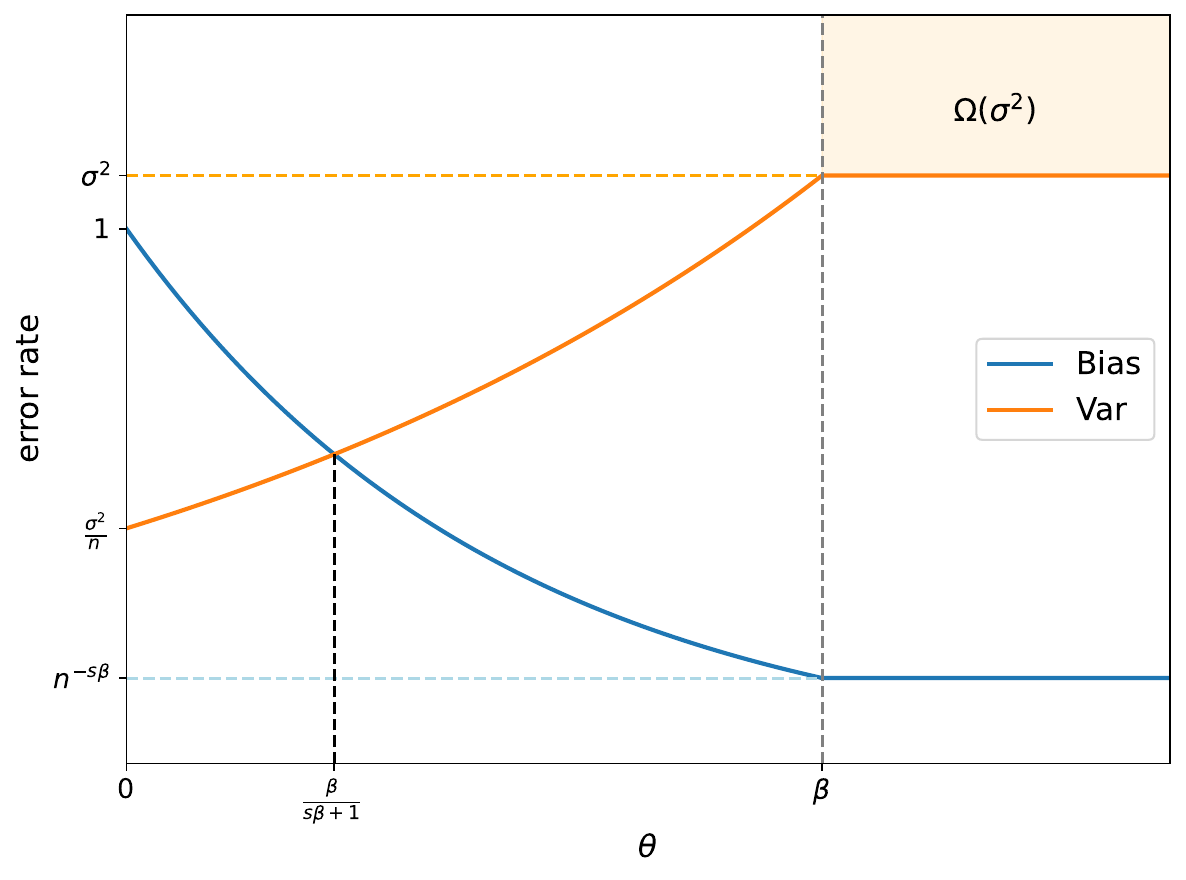}
    \end{minipage}%
  }%
  \subfigure{
    \begin{minipage}[t]{0.5\linewidth}
      \centering
      \includegraphics[width=1\linewidth]{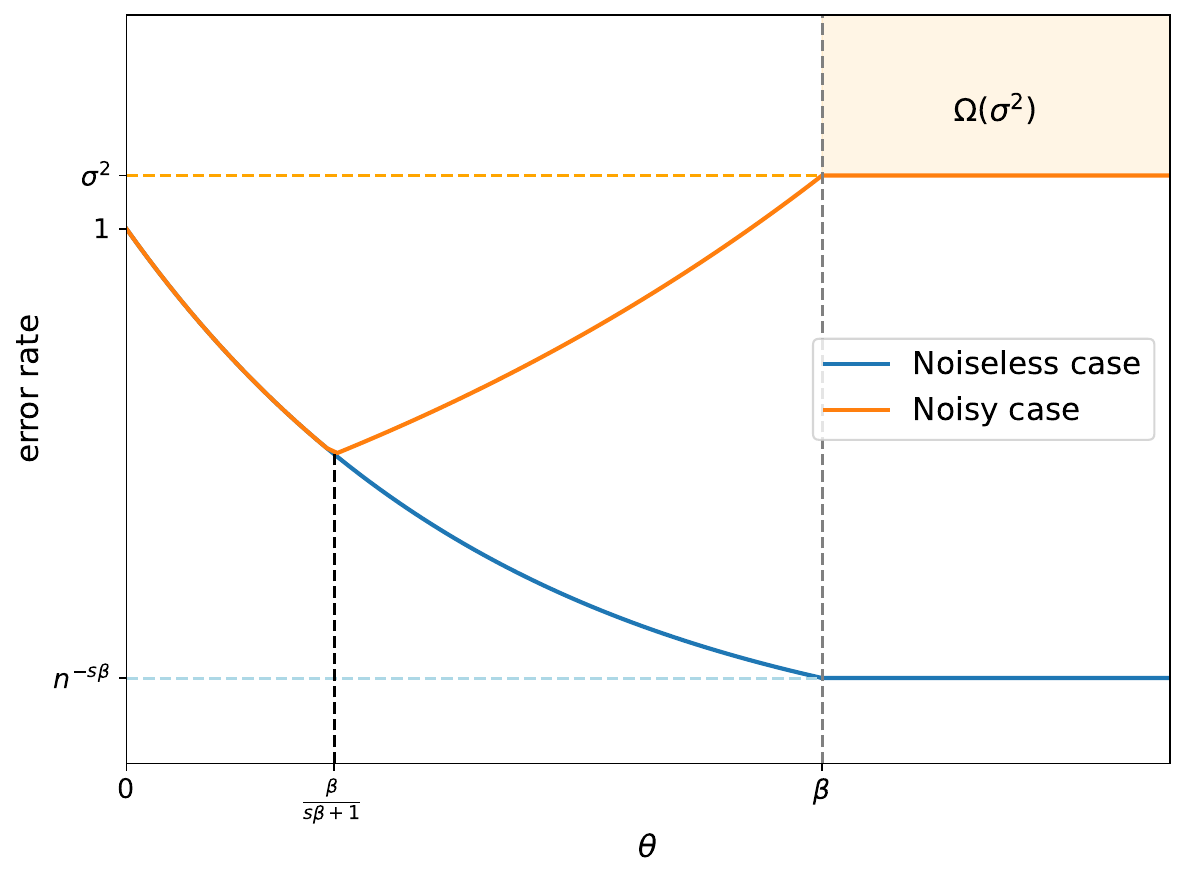}
    \end{minipage}%
  }%

  \subfigure{
    \begin{minipage}[t]{0.5\linewidth}
      \centering
      \includegraphics[width=1\linewidth]{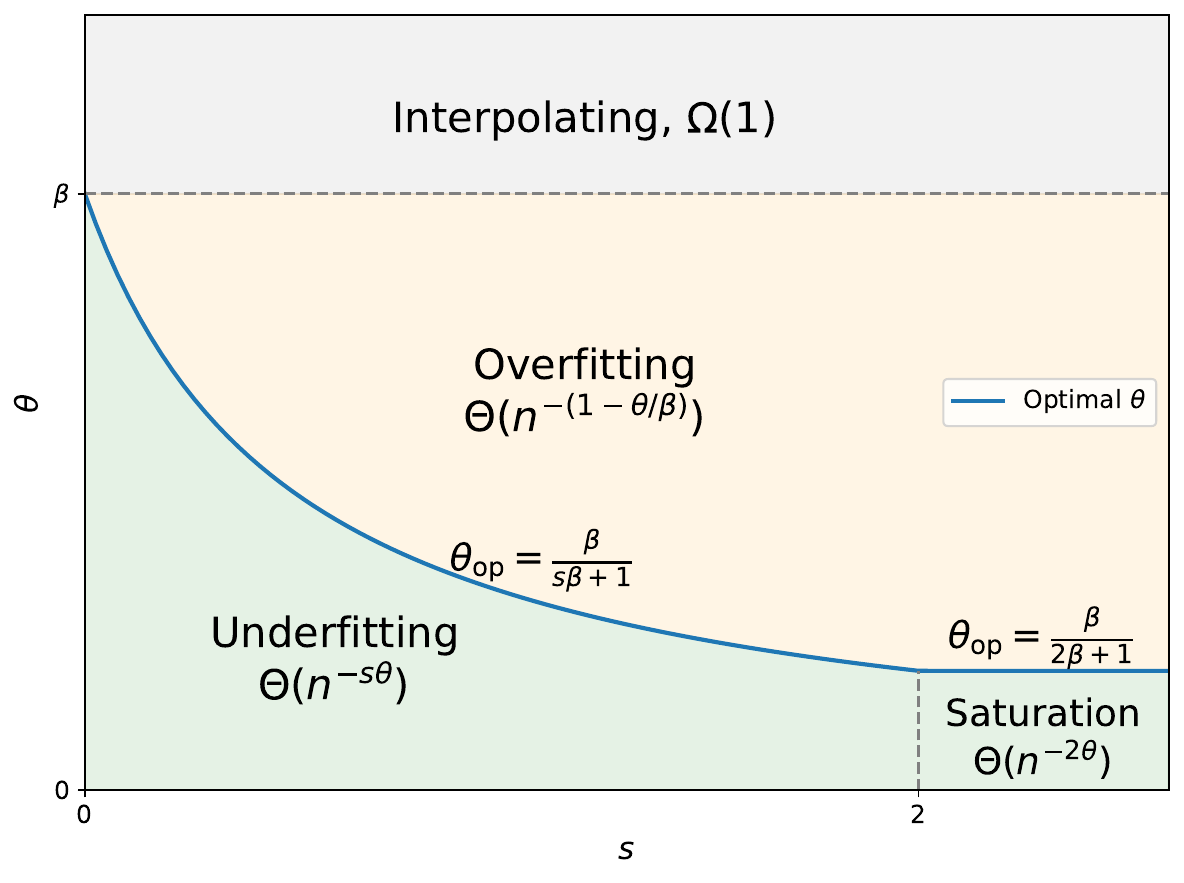}
    \end{minipage}%
  }%
  \subfigure{
    \begin{minipage}[t]{0.5\linewidth}
      \centering
      \includegraphics[width=1\linewidth]{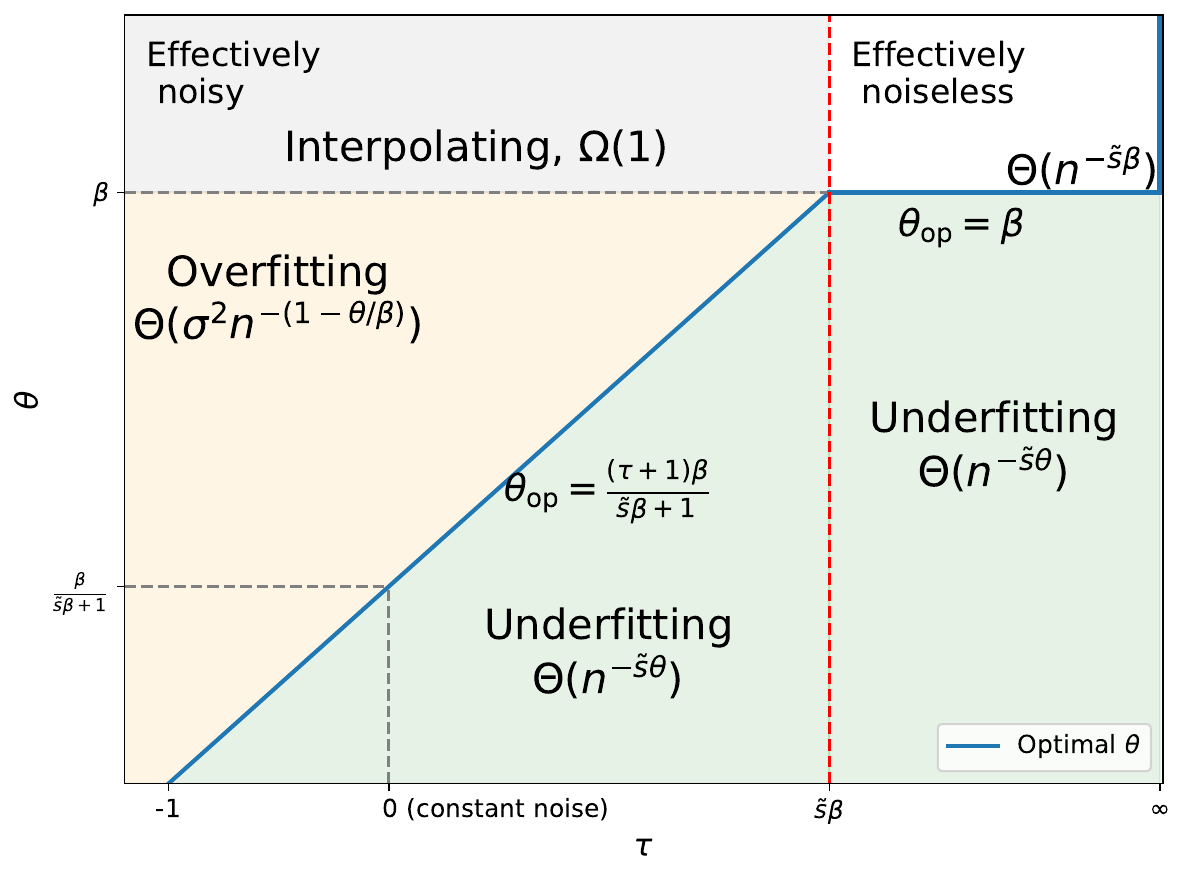}
    \end{minipage}%
  }%

  \centering
  \caption{An illustration of the learning curves when choosing $\lambda = n^{-\theta}$.
  First row: The bias-variance plot and the error curves for the noisy and noiseless cases.
  Second row: Tow phase diagrams of the asymptotic rates of the excess risk with respect to parameter pairs $(\theta, s)$ and $(\theta,\tau)$,
    where we set $\sigma^2 = n^{-\tau}$ and $\tilde{s} = \min(s,2)$.
    In the ``underfitting'' (``overfitting'') region, bias (variance) is dominating.
    The ``interpolating'' region refers to the extreme cases of overfitting that the excess risk is lower bounded by a constant.
    For the first diagram we consider the case of constant noise.
    For the second diagram, the red vertical line shows the crossover of the noisy regime to the noiseless regime
    and an upper bound for the blank area on the upper-right corner is unknown yet.
  }

  \label{fig:LearningCurve}
\end{figure}

Before presenting our main results, we have to introduce a source condition on the regression function.
Since we will establish both precise learning rates,
we have to characterize the exact smoothness order of $f^*_\rho$ rather than merely assume $f^*_\rho$ belongs to some interpolation space $[\caH]^s$.

\begin{assumption}[Source condition]
  \label{assu:source}
  There are some $s > 0$ and a sequence $(a_i)_{i\geq 1}$ such that
  \begin{align}
    \label{eq:fHs}
    f^*_\rho = \sum_{i = 1}^\infty a_{i}\lambda_i^{s/2} i^{-1/2} e_i
  \end{align}
  and $0 < c \leq \abs{a_i} \leq C$ for some constants $c,C$.
\end{assumption}
\begin{remark}
  \cref{assu:source} is also considered in \citet[Eq. (8)]{cui2021_GeneralizationError} and a slightly weaker version of it is given in \citet[Assumption 5]{jin2021_LearningCurves}.
  We only consider this simple form since there is no essential difference in the proof to consider the weaker version.
  From the definition \cref{eq:InterpolationSpaceDef} we can see that \cref{assu:source} implies  $f^*_\rho \in [\caH]^{t}$ for any $t < s$ but $f^*_\rho \notin [\caH]^s$.
\end{remark}


\subsection{Noisy case}

Let us first consider the noisy case with the following assumption:

\begin{assumption}[Noise]
  \label{assu:noise}
  We assume
  \begin{align}
    \E_{(x,y)\sim \rho} \left[ \left( y-f^{*}_{\rho}(x) \right)^2 \;\Big|\; x \right] = \sigma^2 > 0,
    \quad \mu\text{-a.e. } x \in \caX.
  \end{align}
\end{assumption}

For technical reason, we have to further assume the kernel to be Hölder-continuous, which is first in introduced in \citet{li2023_SaturationEffect}.
This assumption is satisfied for the Laplace kernel, Sobolev kernels and neural tangent kernels.
\begin{assumption}
  \label{assu:Holder}
  The kernel $k$ is Hölder-continuous,
  that is, there exists some $p \in (0,1]$ and $L > 0$ such that
  \begin{align}
    \label{eq:Holder}
    \abs{k(x_1,x_2) - k(y_1,y_2)} \leq L \norm{(x_1,x_2) - (y_1,y_2)}_{\R^{d \times d}}^p, \quad
    \forall x_1,x_2,y_1,y_2 \in \caX.
  \end{align}
\end{assumption}

\begin{theorem}
  \label{thm:LearningCurve_Noisy}
  Under Assumptions~\ref{assu:EDR}-\ref{assu:Holder}, suppose $\lambda \asymp n^{-\theta}$ for $\theta > 0$.
  Then,
  \begin{align}
    \E\left[  \norm{\hat{f}_\lambda - f^*_\rho}^2_{L^{2}} \;\Big|\; X \right] =
    \begin{cases}
      \tilde{\Theta}_{\bbP}\mleft( n^{-\min(s,2)\theta} + \sigma^2 n^{-(1 - \theta / \beta)}  \mright), & \qif \theta < \beta \\
      \Omega^{\poly}_{\bbP}\mleft( \sigma^2 \mright), & \qif \theta \geq \beta,
    \end{cases}
  \end{align}
  where $\tilde{\Theta}_{\bbP}$ can be replaced with $\Theta_\bbP$ for the first case if $s \neq 2$.
\end{theorem}

\begin{remark}
  The two terms in the first case in \cref{thm:LearningCurve_Noisy} actually correspond to the bias and the variance term respectively.
  Balancing the two terms, we find the optimal regularization is
  $\theta_{\mathrm{op}} = \frac{\beta}{\tilde{s}\beta+1}$ and the optimal rate is $\frac{\tilde{s}\beta}{\tilde{s}\beta+1}$,
  where $\tilde{s} = \min(s,2)$, which recovers the classical optimal rate results~\citep{caponnetto2007_OptimalRates}.
  Moreover, while we treat $\sigma^2$ as fixed for simplicity, we can also allow $\sigma^2$ to vary with $n$.
  Then, we can recover the results in \citet{cui2021_GeneralizationError}.
\end{remark}

\subsection{Noiseless case}


\begin{theorem}
  \label{thm:LearningCurve_Noiseless}
  Under Assumptions~\ref{assu:EDR}-\ref{assu:source}, assume further that the noise is zero, i.e., $y = f^*_\rho(x)$.
  Then, we have:
  \begin{itemize}
    \item Suppose $\lambda \asymp n^{-\theta}$ for $\theta \in (0,\beta)$, we have
    \begin{align}
      \E\left[  \norm{\hat{f}_\lambda - f^*_\rho}^2_{L^{2}} \;\Big|\; X \right] =
      \tilde{\Theta}_{\bbP}\mleft( n^{-\min(s,2)\theta} \mright),
    \end{align}
    where $\tilde{\Theta}_{\bbP}$ can be replaced with $\Theta_\bbP$ if $s \neq 2$.
    \item Suppose $\lambda \asymp n^{-\theta}$ for $\theta \geq \beta$ and assume further that $s >1$.
    Then,
    \begin{align}
      \E\left[  \norm{\hat{f}_\lambda - f^*_\rho}^2_{L^{2}} \;\Big|\; X \right] =
      O_{\bbP}^{\poly}\mleft( n^{-\min(s,2)\beta} \mright).
    \end{align}
    Moreover, we have the information-theoretical lower rate:
    \begin{align}
      \sup_{\norm{f^*_\rho}_{[\caH]^s} \leq R} \E\left[  \norm{\hat{f}_\lambda - f^*_\rho}^2_{L^{2}} \;\Big|\; X \right]
      = \Omega(n^{-s\beta}),
    \end{align}
    where $R >0$ is a fixed constant.
  \end{itemize}
\end{theorem}

\begin{remark}
  \cref{thm:LearningCurve_Noiseless} shows that the generalization error of KRR in the noiseless case is monotone decreasing when $\theta$ increases
  and reaches the optimal rate $n^{-\beta}$ when $\theta \geq \beta$ if $s \leq 2$.
  Since the case $\theta \to \infty$ corresponds to kernel interpolation, our result implies that kernel interpolation is optimal when there is no noise.
  In contrast, as shown in \cref{thm:LearningCurve_Noisy} (or \citet{li2023_KernelInterpolation}), kernel interpolation can not generalize in the noisy case.
  For the case $s > 2$, the KRR method suffers from saturation and the resulting convergence rate is limited to $n^{-2\beta}$,
  while the possible lower rate is $n^{-s\beta}$.
\end{remark}

\subsection{Discussion}\label{subsec:Discussion}

Our results provide a full picture of the generalization of KRR,
which is in accordance with our traditional belief of the bias-variance trade-off principle:
the generalization error is a U-shaped curve with respect to the regularization parameter $\lambda$ in the noisy case
and is monotone decreasing in the noiseless case.
See \cref{fig:LearningCurve} for an illustration.

Our rates coincide with the upper rates in the traditional KRR literature~\citep{caponnetto2007_OptimalRates,fischer2020_SobolevNorm}.
Moreover, our results also recover the learning curves in \citet{cui2021_GeneralizationError},
but we do not need the strong assumption of Gaussian design eigenfunctions as in \citet{cui2021_GeneralizationError},
which may not be true in most cases.
Our assumptions are mild and hold for a large class of kernels including the Sobolev kernels and the neural tangent kernels (NTK) on spheres.

Our results are based on the bias-variance decomposition and determining the rates for each term respectively.
In the proof of \citet{li2023_SaturationEffect}, they determined the rates of the variance term under the condition that $\theta < \frac{1}{2}$
and that of the bias term when $s \geq 2$ and $\theta < 1$.
The subsequent work \citet{li2023_KernelInterpolation} proved the rates of the variance term when $\theta < \beta$
and provided a near constant lower bound for $\theta \geq \beta$.
Considering the counterpart, our works further prove the rates of the bias term, which finally enables us to determine the complete
learning curve of KRR\@.

The connection between KRR and Gaussian process regression also results in the connection between their learning curves.
\citet{jin2021_LearningCurves} claimed to show learning curves for Gaussian process regression.
However, regardless of the gap in their proof as pointed out in \citet{zhang2023_OptimalityMisspecifieda}, their results are more restrictive than ours.
Considering a boundedness assumption of the eigenfunctions that $\norm{e_i}_{\infty} \leq Ci^{\tau}$ for some $\tau \geq 0$,
they could only cover the regime of $\theta < \beta/(1+2\tau)$.
Moreover, to approach the $\theta = \beta$ regime for the $\Omega(1)$ bound in the noisy case or the optimal rate in noiseless case,
they have to require $\tau=0$, that is, the eigenfunctions are uniformly bounded,
but it is not true for some kernels such as dot-product kernels on spheres (and thus for NTK) since in general spherical harmonics are not uniformly bounded.
In contrast, our embedding index assumption still holds in this case.

  \section{Proof sketch}\label{sec:proof-sketch}
  We first introduce the following sample versions of the auxiliary integral operators, which are commonly used in the related literature
~\citep{caponnetto2007_OptimalRates,fischer2020_SobolevNorm,li2023_SaturationEffect}.
We define the sampling operator $K_{x} : \R \to \caH$ by $K_x y = yk(x,\cdot)$,
whose adjoint $K_x^*: \caH \to \R$ is given by $K_x^* f = f(x)$.
The sample covariance operator $T_X : \caH \to \caH$ is defined by
\begin{align}
  \label{eq:TX}
  T_X &\coloneqq \frac{1}{n}\sum_{i=1}^n K_{x_i} K_{x_i}^*,
\end{align}
and the sample basis function is $g_Z  \coloneqq \frac{1}{n} \sum_{i=1}^n K_{x_i}y_i \in \caH.$
As shown in \citet{caponnetto2007_OptimalRates}, the operator form of KRR writes
\begin{align}
  \label{eq:KRR}
  \hat{f}_\lambda = (T_X+\lambda)^{-1} g_Z.
\end{align}

Let us further define
\begin{align}
  \tilde{g}_{Z} \coloneqq \mathbb{E}\left( g_{Z} | X \right) = \frac{1}{n} \sum_{i=1}^n K_{x_i} f_{\rho}^{*}(x_{i}) \in \mathcal{H},
\end{align}
and
\begin{align}
  \tilde{f}_{\lambda} \coloneqq \mathbb{E}\left( \hat{f}_{\lambda} | X \right) = \left(T_{X} + \lambda\right)^{-1} \tilde{g}_{Z} \in \mathcal{H}.
\end{align}
Then, the traditional bias-variance decomposition~\citep{li2023_SaturationEffect,zhang2023_OptimalityMisspecified} yields
\begin{align}
  \label{eq:4_BVDecomp}
  \E \left( \norm{\hat{f}_\lambda - f^*_\rho}^2_{L^2} \;\Big|\; X \right)
  =  \mathbf{Bias}^2(\lambda) + \mathbf{Var}(\lambda),
\end{align}
where
\begin{equation}
  \label{eq:4_BVFromula}
  \begin{aligned}
    & \mathbf{Bias}^2(\lambda) \coloneqq
    \norm{\tilde{f}_{\lambda} - f_{\rho}^{*}}_{L^2}^2, \quad
    \mathbf{Var}(\lambda)\coloneqq
    \frac{\sigma^2}{n^2} \sum_{i=1}^n  \norm{(T_X+\lambda)^{-1} k(x_i,\cdot)}^2_{L^2}.
  \end{aligned}
\end{equation}

\subsection{The noisy case}
To prove the desired result, we have to establish the asymptotic orders of both $\mathbf{Bias}^2(\lambda)$ and $\mathbf{Var}(\lambda)$.
We first prove the asymptotic order of $\mathbf{Bias}^2(\lambda)$ as one of our technical contributions.
As far as we know, we are the first to provide such a lower bound in \cref{eq:BiasAsymp}.
\begin{lemma}
  \label{lem:BiasAsymp}
  Under Assumptions~\ref{assu:EDR},\ref{assu:EMB},\ref{assu:source}, suppose $\lambda \asymp n^{-\theta}$ for $\theta \in (0,\beta)$.
  Then,
  \begin{align}
    \label{eq:BiasAsymp}
    \mathbf{Bias}^2(\lambda) = \tilde{\Theta}_{\bbP}\mleft( n^{-\min(s,2)\theta} \mright),
  \end{align}
  where $\tilde{\Theta}_{\bbP}$ can be replaced with $\Theta_\bbP$ if $s \neq 2$.
\end{lemma}
\begin{proof}
  [Proof sketch of \cref{lem:BiasAsymp}]
  Denote $\tilde{s} = \min(s,2)$.
  We first introduce the regularized regression function $f_\lambda \coloneqq T(T+\lambda)^{-1}f^*_\rho$
  and triangle inequality implies
  \begin{align*}
    \mathbf{Bias}(\lambda) = \norm{\tilde{f}_{\lambda} - f_{\rho}^*}_{L^2} \geq \norm{f_\lambda - f_{\rho}^*}_{L^2} - \norm{\tilde{f}_{\lambda} - f_\lambda}_{L^2}.
  \end{align*}
  There is no randomness in the first term and we can use the expansion \cref{eq:fHs} and \cref{eq:Mercer} to show that
  $\norm{f_\lambda - f_{\rho}^*}_{L^2} = \tilde{\Theta}\left( n^{-\tilde{s}\theta} \right)$.
  Then, we have to prove the error term $\norm{\tilde{f}_{\lambda} - f_\lambda}_{L^2}$ to be infinitesimal with respect to the main term,
  which is the main difficulty since it requires a refined analysis.
  Previous work only consider the case $\theta = \frac{\beta}{\tilde{s}\beta + 1}$ (corresponding to the optimal regularization) and show an
  $O(n^{-\tilde{s}\theta})$ bound rather than the $o(n^{-\tilde{s}\theta})$ bound that we require.
  For the proof, we
  (1) apply the concentration techniques in \citet{fischer2020_SobolevNorm};
  (2) consider the $L^q$-embedding property in \citet{zhang2023_OptimalityMisspecified} for the mis-specified case when $s$ is small;
  (3) sharpen the estimation by exploiting the embedding property $\alpha_0 = 1/\beta$ and $\theta < \beta$.
  For the detail, see Section 2.2 in the supplementary material.
\end{proof}

The variance term has been analyzed in \citet{li2023_KernelInterpolation}.
We present the following proposition as a combination of Proposition 5.3 and Theorem 5.10 in \citet{li2023_KernelInterpolation}.
\begin{proposition}
  \label{prop:VarAsymp}
  Under Assumptions~\ref{assu:EDR}-\ref{assu:Holder}, suppose that $\lambda \asymp n^{-\theta}$.
  Then,
  \begin{align}
    \mathbf{Var}(\lambda) =
    \begin{cases}
      \Theta_{\bbP}^{\poly}\mleft( \sigma^2 n^{-(1-\theta/\beta)} \mright), & \qif \theta < \beta; \\
      \Omega_{\bbP}^{\poly}\mleft( \sigma^2 \mright), & \qif \theta \geq \beta.
    \end{cases}
  \end{align}
\end{proposition}

\subsection{The noiseless case}

For the noiseless case, the variance term vanishes in \cref{eq:4_BVDecomp}, and thus we only need to consider the bias term.
Since we have already established the estimation for large $\lambda$ in \cref{lem:BiasAsymp}, we focus on the case of small $\lambda$.
\begin{lemma}
  \label{lem:BiasUpperInterpolating}
  Under Assumptions~\ref{assu:EDR},\ref{assu:EMB},\ref{assu:source}, assume further $s > 1$.
  Suppose $\lambda \asymp n^{-\theta}$ for $\theta \geq \beta$.
  Then,
  \begin{align}
    \mathbf{Bias}^2(\lambda) = O^{\poly}_{\bbP}(n^{-\min(s,2)\beta}).
  \end{align}
\end{lemma}
\begin{proof}
  [Proof sketch of \cref{lem:BiasUpperInterpolating}]
  Intuitively, we hope to bound $\mathbf{Bias}^2(\lambda)$ with $\mathbf{Bias}^2(\tilde{\lambda})$ for $\tilde{\lambda} > \lambda$ such that concentration still works.
  However, we can not directly derive no monotone property of $\mathbf{Bias}(\lambda)$.
  Nevertheless, since $f^*_\rho \in \caH$ when $s > 1$, the bias term can be written as
  \begin{align*}
      \mathbf{Bias}(\lambda) =  \norm{\lambda (T_X+\lambda)^{-1}f^{*}_{\rho}}_{L^2} = \norm{T^{\frac{1}{2}}\lambda (T_X+\lambda)^{-1}f^{*}_{\rho}}_{\caH}
      \leq \norm{T^{\frac{1}{2}}\lambda (T_X+\lambda)^{-1}}_{\mathscr{B}(\caH)} \norm{f^{*}_{\rho}}_{\caH}.
  \end{align*}
  Then, by operator calculus we can show that
  \begin{align*}
      \norm{T^s \left[ \lambda (T_X+\lambda)^{-1} \right]}_{\mathscr{B}(\caH)} \leq
      \norm{T^s \left[ \tilde{\lambda} (T_X+\tilde{\lambda})^{-1} \right]}_{\mathscr{B}(\caH)}
  \end{align*}
  reducing $\lambda$ to $\tilde{\lambda}$.
  Now, we can replace $T_X$ with $T$ using concentration results and derive the desired upper bound.
\end{proof}

The following proposition shows that the upper bound in \cref{lem:BiasUpperInterpolating} matches the information-theoretical lower bound.
The proof follows idea of the minimax principle~\citep{micchelli1979_DesignProblems} and is deferred to the supplementary material.
\begin{proposition}
  \label{prop:BiasLower}
  Suppose \cref{assu:EDR} holds and $s \geq 1$.
  For any $X = (x_1,\dots,x_n)$, we have
  \begin{align}
    \sup_{\norm{f^*_\rho}_{[\caH]^s} \leq R} \mathbf{Bias}^2(\lambda) = \Omega\mleft(n^{-s\beta}\mright),
  \end{align}
  where we note that here $\mathbf{Bias}(\lambda)$ is viewed as a function depending also on $f^*_\rho$ and $X$.

\end{proposition}

  \section{Experiments}\label{sec:experiments}
  Lots of numerical experiments on both synthetic data and real data are done to study to learning curves of KRR
~\citep{li2023_SaturationEffect,cui2021_GeneralizationError}.
In this section, we consider numerical experiments on a toy model to verify our theory.

Let us consider the kernel $k(x,y) = \min(x,y)$ and $x  \sim \mathcal{U}[0,1]$.
Then, the corresponding RKHS is~\citep{wainwright2019_HighdimensionalStatistics}
\begin{align*}
  \caH =  \left\{ f : [0,1] \to \R ~\Big|~ f \text{ is absolutely continuous, } f(0) = 0,\ \int_0^1 (f'(x))^2 \dd x < \infty \right\}
\end{align*}
and the eigenvalue decay rate $\beta = 2$.
Moreover, the eigensystem of $k$ is known to be $\lambda_i = \left( \frac{2i-1}{2}\pi \right)^{-2}$
and $e_i(x) = \sqrt{2} \sin(\frac{2i-1}{2}\pi x)$, which allows us to directly compute the smoothness of certain functions.
For some $f^*$, we generate data from the model $y = f^*(x) + \ep$ where $\ep \sim \mathcal{N}(0,0.05)$
and perform KRR with $\lambda = c n^{-\theta}$ for different $\theta$'s with some fixed constant $c$.
Then, we numerically compute the variance, bias and excess risk by Simpson's formula with $N \gg n$ nodes.
Repeating the experiment for $n$ ranged in 1000 to 5000, we can estimate the convergence rate $r$ by a logarithmic least-squares
$\log \text{err} = r \log n + b$ on the values (variance, bias and excess risk).
The results are collected in \cref{tab:Table1}.
It can be seen that the resulting values basically match the theoretical values and we conclude that our theory is supported by the experiments.
For more experiments and more details, we refer to the supplementary material.

\begin{table}[ht]
  \small
  \renewcommand\arraystretch{1.5}
  \centering
  \begin{tabular}{|c|c|cc|cc|cc|}
    \hline
    & $f^*(x)=$ & \multicolumn{2}{c|}{$\cos{2 \pi x}$ ($s=\frac{1}{2}$)} & \multicolumn{2}{c|}{$\sin{2 \pi x}$ ($s=1.5$)} & \multicolumn{2}{c|}{$\sin{\frac{3}{2} \pi x}$ ($s =\infty$)}   \\
    \hline
    $\theta$ & Variance    & Bias        & Risk                 & Bias        & Risk                 & Bias        & Risk                 \\
    \hline
    0.2      & 0.90 (0.90) & 0.13 (0.10) & 0.13 (0.10)          & 0.34 (0.30) & 0.34 (0.30)          & 0.40 (0.40) & 0.42 (0.40)          \\
    \hline
    0.4      & 0.80 (0.80) & 0.22 (0.20) & 0.22 (0.20)          & 0.68 (0.60) & 0.69 (0.60)          & 0.82 (0.80) & \textbf{0.81 (0.80)} \\
    \hline
    0.5      & 0.75 (0.75) & 0.26 (0.25) & 0.26 (0.25)          & 0.84 (0.75) & \textbf{0.79 (0.75)} & 1.04 (1.00) & 0.77 (0.75)  \\
    \hline
    1.0      & 0.49 (0.50) & 0.54 (0.50) & \textbf{0.52 (0.50)} & 1.69 (1.50) & 0.49 (0.50)          & 2.21 (2.00) & 0.49 (0.50)  \\
    \hline
    2.0      & 0.00 (0.00) & 1.05 (1.00) & 0.09 (0.00)          & 3.26 (3.00) & 0.00 (0.00)          & 3.99 (4.00) & 0.00 (0.00)          \\
    \hline
    3.0      & 0.00 (0.00) & 1.05 (1.00) & 0.09 (0.00)          & 3.26 (3.00) & 0.00 (0.00)          & 3.98 (4.00) & 0.00 (0.00)          \\

    \hline
  \end{tabular}
  \caption{Asymptotic rates of bias, variance and excess risk under three regressions and different choices of $\theta$.
  The numbers in parenthess are the theoretical values.
  The bolded cells correspond to the best rate over the choices of $\theta$'s.}
  \label{tab:Table1}
\end{table}

  \section{Conclusion}
  In this paper, we prove rigorously the learning curves of KRR,
showing the interplay of the eigenvalue decay of the kernel, the relative smoothness of the regression function,
the noise and the choice of the regularization parameter.
The results justify our traditional bias-variance trade-off principle and
provide a full picture of the generalization performance of KRR\@.
These results will help us better understand the generalization mystery of neural networks.

As for future works, we notice that for the nearly interpolating regime when $\theta \geq \beta$,
there are still some missing parts due to technical limitations.
We expect that further analysis will prove the exact orders of the variance term like
that given in \citet{mallinar2022_BenignTempered} under the Gaussian design assumption.
We also hypothesize that \cref{lem:BiasUpperInterpolating} still holds in the mis-specified case ($s < 1$).

\begin{ack}
This work is supported in part by the Beijing Natural Science Foundation (Grant Z190001)
and National Natural Science Foundation of China (Grant 11971257).


\end{ack}

  \clearpage
  \appendix

  \section{Detailed proofs}\label{sec:supplementary-material}

The first step of the proof is the traditional bias-variance decomposition.
Let us further define
\begin{align}
  \tilde{g}_{Z} \coloneqq \mathbb{E}\left( g_{Z} | X \right) = \frac{1}{n} \sum_{i=1}^n K_{x_i} f_{\rho}^{*}(x_{i}) \in \mathcal{H},
\end{align}
and
\begin{align}
  \tilde{f}_{\lambda} \coloneqq \mathbb{E}\left( \hat{f}_{\lambda} | X \right) = \left(T_{X} + \lambda\right)^{-1} \tilde{g}_{Z} \in \mathcal{H}.
\end{align}
Recalling \cref{eq:KRR}, we have
\begin{align*}
  \hat{f}_{\lambda}
  &= \frac{1}{n} (T_X+\lambda)^{-1} \sum_{i=1}^n K_{x_i}y_i = \frac{1}{n} (T_X+\lambda)^{-1} \sum_{i=1}^n K_{x_i} (f^{*}_{\rho}(x_i) + \epsilon_i) \\
  &= (T_X+\lambda)^{-1} \tilde{g}_Z + \frac{1}{n}\sum_{i=1}^n (T_X+\lambda)^{-1} K_{x_i}\epsilon_i,
\end{align*}
so that
\begin{align*}
  \hat{f}_{\lambda}  - f^{*}_{\rho} = \left( \tilde{f}_{\lambda} - f_{\rho}^{*} \right) + \frac{1}{n}\sum_{i=1}^n (T_X+\lambda)^{-1} K_{x_i}\epsilon_i.
\end{align*}
Taking expectation over the noise $\epsilon$ conditioned on $X$, since $\ep | x$ are independent noise with mean 0 and variance $\sigma^{2}$, we have
\begin{align}
  \label{eq:BiasVarianceDecomp}
  \E \left( \norm{\hat{f}_\lambda - f^{*}_{\rho}}^2_{L^2} \;\Big|\; X \right)
  =  \mathbf{Bias}^2(\lambda) + \mathbf{Var}(\lambda),
\end{align}
where
\begin{equation}
  \label{eq:a_BiasVarFormula}
  \begin{aligned}
    & \mathbf{Bias}^2(\lambda) \coloneqq
    \norm{\tilde{f}_{\lambda} - f_{\rho}^{*}}_{L^2}^2, \quad
    \mathbf{Var}(\lambda)\coloneqq
    \frac{\sigma^2}{n^2} \sum_{i=1}^n  \norm{(T_X+\lambda)^{-1} k(x_i,\cdot)}^2_{L^2}.
  \end{aligned}
\end{equation}

\subsection{The variance term}

\begin{theorem}
  \label{thm:VarAsymp}
  Under Assumptions~\ref{assu:EDR}-\ref{assu:Holder}, suppose that $\lambda \asymp n^{-\theta}$.
  Then,
  \begin{align}
    \mathbf{Var}(\lambda) =
    \begin{cases}
      \Theta_{\bbP}^{\poly}\mleft( \sigma^2 n^{-(1-\theta/\beta)} \mright), & \qif \theta < \beta; \\
      \Omega_{\bbP}^{\poly}\mleft( \sigma^2 \mright), & \qif \theta \geq \beta.
    \end{cases}
  \end{align}
\end{theorem}
The computation in \citet{li2023_SaturationEffect} shows that
\begin{align*}
  \mathbf{Var}(\lambda) = \frac{\sigma^2}{n^2} \int_{\caX} \K(x,X)(K+\lambda)^{-2}\K(X,x) \dd \mu(x).
\end{align*}
Then, \cref{thm:VarAsymp} directly follows from Proposition 5.3 and Theorem 5.10 in \citet{li2023_KernelInterpolation}.

\subsection{The bias term}

\begin{theorem}
  \label{thm:BiasAsymp}
  Under Assumptions~\ref{assu:EDR},\ref{assu:EMB},\ref{assu:source}, suppose $\lambda \asymp n^{-\theta}$ for $\theta \in (0,\beta)$.
  Then,
  \begin{align}
    \mathbf{Bias}^2(\lambda) = \tilde{\Theta}_{\bbP}\left( n^{-\min(s,2)\theta} \right),
  \end{align}
  where $\tilde{\Theta}_{\bbP}$ can be replaced with $\Theta_\bbP$ if $s \neq 2$.
\end{theorem}

Let us define the regularized version of the regression function
\begin{align}
  f_{\lambda} \coloneqq (T+\lambda)^{-1}Tf^{*}_{\rho}.
\end{align}
Then, the triangle inequality implies that
\begin{align}
  \label{eq:a_BiasTriangle}
  \mathbf{Bias}(\lambda) = \norm{\tilde{f}_{\lambda} - f_{\rho}^*}_{L^2} \geq
  \norm{f_\lambda - f_{\rho}^*}_{L^2} - \norm{\tilde{f}_{\lambda} - f_\lambda}_{L^2}
\end{align}

Then, the proof of \cref{thm:BiasAsymp} is the combination of the following \cref{lem:Approximation_Main} (with $\gamma = 0$) and \cref{lem:Approximation_Error},
showing that the main term $\norm{f_\lambda - f_{\rho}^*}_{L^2} = \tilde{\Theta}_{\bbP}\left( n^{-\min(s,2)\theta/2} \right)$
and the error term $\norm{\tilde{f}_{\lambda} - f_\lambda}_{L^2} = o_{\bbP}\left( n^{-\min(s,2)\theta/2} \right)$.

\begin{lemma}
  \label{lem:Approximation_Main}
  Under Assumptions~\ref{assu:EDR} and~\ref{assu:source},
  for any $0 \leq \gamma < s$, we have
  \begin{equation}
    \label{eq:Approximation}
    \norm{f_\lambda - f_{\rho}^*}_{[\mathcal{H}]^{\gamma}}^2 \asymp
    \begin{cases}
      \lambda^{s-\gamma}, & s -\gamma < 2; \\
      \lambda^{2}\ln \frac{1}{\lambda}, &  s -\gamma = 2; \\
      \lambda^{2}, & s -\gamma > 2.
    \end{cases}
  \end{equation}
\end{lemma}
\begin{proof}
  From the definition of interpolating norms, letting $p = (s-\gamma)/2$, we have
  \begin{align}
    \norm{f_\lambda - f_{\rho}^*}_{[\mathcal{H}]^{\gamma}}^2
    &= \sum_{i = 1}^\infty  a_i^2 \frac{\lambda^2 }{(\lambda_i + \lambda)^2} (\lambda_i^{s} i^{-1}) \lambda_i^{-\gamma}
    \asymp \lambda^2 \sum_{i = 1}^\infty  \left( \frac{\lambda_i^{p} }{\lambda_i + \lambda} \right)^2 i^{-1}.
  \end{align}
  Then result then follows by applying \cref{prop:SeriesAsymp} for the last series.
\end{proof}

The following lemma shows the error term in \cref{eq:a_BiasTriangle} is infinitesimal, whose proof relies on fine-grained
concentration results established in \cref{subsec:a_approximations}.

\begin{lemma}
  \label{lem:Approximation_Error}
  Under Assumptions~\ref{assu:EDR}-\ref{assu:source}.
  Suppose $\lambda \asymp n^{-\theta}$ with $\theta \in (0,\beta)$, then
  \begin{align}
    \norm{\tilde{f}_{\lambda} - f_{\lambda}}_{L^2} = o_{\bbP}\left( n^{-\min(s,2)\theta/2} \right)
  \end{align}
\end{lemma}
\begin{proof}
  We begin with
  \begin{align}
    \notag
    \norm{\tilde{f}_\lambda - f_{\lambda}}_{L^2} &= \norm{ T^{\frac{1}{2}} \left( \tilde{f}_\lambda-f_{\lambda} \right)}_{\caH} \\
    \label{eq:a_ApproxErrorDecomp}
    &\leq \norm{T^{\frac{1}{2}} T_{\lambda}^{-\frac{1}{2}}} \cdot \norm{ T_{\lambda}^{\frac{1}{2}} T_{X \lambda}^{-1} T_{\lambda}^{\frac{1}{2}} }
    \cdot \norm{T_{\lambda}^{-\frac{1}{2}} \left( \tilde{g}_{Z} - T_{X \lambda} f_{\lambda} \right)}_{\caH}.
  \end{align}
  From operator calculus we know $\norm{T^{\frac{1}{2}} T_{\lambda}^{-\frac{1}{2}}} \leq 1$.
  Moreover, since $\theta < \beta$ and the embedding index $\alpha_0 = 1/\beta$, by \cref{lem:ConcenIneq} we get
  $\norm{ T_{\lambda}^{\frac{1}{2}} T_{X \lambda}^{-1} T_{\lambda}^{\frac{1}{2}} } \leq 3$
  with high probability as long as $n$ is sufficiently large.
  For the last term in \cref{eq:a_ApproxErrorDecomp}, we have
  \begin{align*}
    T_{\lambda}^{-\frac{1}{2}} \left( \tilde{g}_{Z} - T_{X \lambda} f_{\lambda} \right)
    &= T_\lambda^{-\frac{1}{2}}\left[\left(\tilde{g}_Z - \left(T_X + \lambda + T - T \right) f_\lambda\right)\right] \\
    &= T_\lambda^{-\frac{1}{2}}\left[\left(\tilde{g}_Z - T_X f_\lambda\right) - \left(T + \lambda \right) f_\lambda + T f_\lambda \right] \\
    &= T_\lambda^{-\frac{1}{2}}\left[\left(\tilde{g}_Z-T_X f_\lambda\right)-\left(g-T f_\lambda\right)\right].
  \end{align*}
  Therefore, \cref{lem:a_RegApprox1} and \cref{lem:a_RegApprox2} show that
  \begin{align*}
    \norm{T_{\lambda}^{-\frac{1}{2}} \left( \tilde{g}_{Z} - T_{X \lambda} f_{\lambda} \right)}_{\caH}
    = \norm{T_\lambda^{-\frac{1}{2}}\left[\left(\tilde{g}_Z-T_X f_\lambda\right)-\left(g-T f_\lambda\right)\right]}_{\caH}
    = o_{\bbP}\left( n^{-\min(s,2)\theta/2} \right)
  \end{align*}
  for both $s > \alpha_0$ and $s \leq \alpha_0$ cases.
\end{proof}

\subsection{Approximation results}\label{subsec:a_approximations}

Let us further denote
\begin{align}
  \label{eq:a_Proof_xi}
  \xi(x) =  T_{\lambda}^{-\frac{1}{2}}(K_{x} f_{\rho}^{*}(x) - T_{x} f_{\lambda}).
\end{align}
Then, it is easy to check that
\begin{align*}
  T_\lambda^{-\frac{1}{2}}\left[\left(\tilde{g}_Z-T_X f_\lambda\right)-\left(g-T f_\lambda\right)\right]
  = \frac{1}{n}\sum_{i=1}^n \xi(x_i) - \E_{x \sim \mu} \xi(x).
\end{align*}

The following lemma deals with the easy case when $s > \alpha_0$.
\begin{lemma}
  \label{lem:a_RegApprox1}
  Suppose Assumptions~\ref{assu:EDR}-\ref{assu:source} hold and $s > \alpha_0$.
  Let $\lambda \asymp n^{-\theta}$ with $\theta \in (0,\beta)$ and $\delta \in (0,1)$.
  Then, for $\alpha > \alpha_0 = \beta^{-1}$ being sufficiently close, it holds with probability at least $1-\delta$ that
  \begin{align}
    \norm{T_\lambda^{-\frac{1}{2}}\left[\left(\tilde{g}_Z-T_X f_\lambda\right)-\left(g-T f_\lambda\right)\right]}_{\caH}
    \leq C \ln  \frac{2}{\delta} \cdot \left( M_{\alpha}^2 \frac{ \lambda^{-\alpha}}{n}
    + M_{\alpha} \sqrt {\frac{\lambda^{-\alpha} \ln n}{n}} \right) \lambda^{\tilde{s}/2},
  \end{align}
  where $\tilde{s} = \min(s,2)$.
  Consequently,
  \begin{align}
    \label{eq:a_RegApprox1}
    \norm{T_\lambda^{-\frac{1}{2}}\left[\left(\tilde{g}_Z-T_X f_\lambda\right)-\left(g-T f_\lambda\right)\right]}
    = o_{\bbP}(\lambda^{\tilde{s}/2})
    = o_{\bbP}(n^{-\tilde{s}\theta/2}).
  \end{align}
\end{lemma}

Before proving \cref{lem:a_RegApprox1}, we have to introduce the following proposition bounding the $\gamma$-norms of the regularized basis function,
which is a part of \citet[Corollary 5.6]{li2023_KernelInterpolation}.
\begin{proposition}
  Suppose $\caH$ has embedding index $\alpha_0$.
  Then for any $\alpha > \alpha_0$,
  \begin{align}
    \label{eq:a_RegK_H}
    \norm{T_{\lambda}^{-1/2} k(x,\cdot)}_{\caH} \leq M_\alpha \lambda^{-\alpha/2},\quad \mu\text{-a.e.}~ x\in \caX.
  \end{align}
\end{proposition}

\begin{proof}
[Proof of \cref{lem:a_RegApprox1}]
  To use Bernstein inequality in \cref{lem:ConcenHilbert}, let us bound the $m$-th moment of $\xi(x)$:
  \begin{align}
    \label{eq:a_ProofApprox1_Moment}
    \E \norm{\xi(x)}_{\mathcal{H}}^{m} &= \E \norm{ T_{\lambda}^{-\frac{1}{2}} K_{x}(f_{\rho}^*(x) - f_{\lambda}(x)) }_{\caH}^{m} \notag \\
    &\leq \E \left[ \norm{T_{\lambda}^{-\frac{1}{2}} k(x,\cdot)}_{\mathcal{H}}^{m} \cdot  \E \big( \abs{f_{\rho}^*(x) - f_{\lambda}(x)}^m ~\big|~ x\big) \right].
  \end{align}
  The first term in \cref{eq:a_ProofApprox1_Moment} is bounded through \cref{eq:a_RegK_H}.
  For the second term, since $s > \alpha_0$, using the embedding condition and \cref{lem:Approximation_Main}, we have
  \begin{align*}
    \norm{f_{\lambda} - f_{\rho}^{*}}_{L^\infty} \leq M_{\alpha} \norm{f_{\lambda}  - f_{\rho}^{*}}_{[\caH]^{\alpha}}
    \leq C M_{\alpha} \lambda^{\min(s-\alpha,2)/2} \leq C M_{\alpha} \lambda^{(\tilde{s}-\alpha)/2},
  \end{align*}
  where we notice that $\min(s-\alpha,2) = \min(s,2+\alpha)-\alpha \geq \tilde{s} - \alpha$ for the last inequality.
  Moreover, \cref{lem:Approximation_Main} also implies
  \begin{align*}
    \E \abs{f_{\lambda}(x) - f_{\rho}^{*}(x)}^2 = \norm{f_{\lambda}(x) - f_{\rho}^{*}(x)}_{L^2}^2 \leq C \lambda^{\tilde{s}}\ln\frac{1}{\lambda}
    \leq C \lambda^{\tilde{s}} \ln n.
  \end{align*}
  Plugging in these estimations in \cref{eq:a_ProofApprox1_Moment}, we get
  \begin{align}
    \label{eq:a_ProofApprox1_Bernstein}
    \text{\cref{eq:a_ProofApprox1_Moment}}
    &\leq (M_{\alpha} \lambda^{-\alpha/2})^{m} \cdot  \norm{f_{\lambda} - f_{\rho}^{*}}_{L^{\infty}}^{m-2} \cdot
    \E \abs{f_{\lambda}(x) - f_{\rho}^{*}(x)}^2 \notag \\
    &\le  (M_{\alpha} \lambda^{-\alpha/2})^{m} \cdot \left( C M_{\alpha} \lambda^{(\tilde{s}-\alpha)/2} \right)^{m-2}
    \cdot   (C \lambda^{\tilde{s}} \ln n) \notag \\
    &\le \frac{1}{2} m!
    \left( C M_\alpha^2 \lambda^{\tilde{s}-\alpha} \ln n \right) \cdot \left( C M_{\alpha}^2 \lambda^{-\alpha + \tilde{s}/2} \right)^{m-2}.
  \end{align}
  The proof is then complete by \cref{lem:ConcenHilbert}.
\end{proof}

The case of $s \leq \alpha_0$ is more difficult.
We will use the truncation technique introduced in \citet{zhang2023_OptimalityMisspecified}.
The following lemma can be proven similarly to \cref{lem:Approximation_Main}.

\begin{lemma}
  \label{lem:RegF_Norms}
  Under Assumptions~\ref{assu:EDR} and~\ref{assu:source}, for any $0 \leq \gamma < s+2$, we have
  \begin{equation}
    \label{eq:RegF_Norms}
    \norm{f_\lambda}_{[\mathcal{H}]^{\gamma}}^2 \asymp
    \begin{cases}
      \lambda^{s-\gamma}, & s < \gamma; \\
      \ln \frac{1}{\lambda}, &  s = \gamma; \\
      1, & s > \gamma.
    \end{cases}
  \end{equation}
\end{lemma}
\begin{proof}
  Simply notice that
  \begin{align*}
    \norm{f_\lambda}_{[\mathcal{H}]^{\gamma}}^2
    &= \sum_{i = 1}^\infty  a_i^2 \frac{\lambda_i^2 }{(\lambda_i + \lambda)^2} (\lambda_i^{s} i^{-1}) \lambda_i^{-\gamma}
    \asymp  \sum_{i = 1}^\infty  \left( \frac{\lambda_i^p }{\lambda_i + \lambda} \right)^2 i^{-1},
  \end{align*}
  where $p = (s+2-\gamma)/2$.
  Then we can apply \cref{prop:SeriesAsymp}.
\end{proof}

Then, we are able to show the following concentration result about the truncated $\xi_i$'s,
whose proof resembles that of \cref{lem:a_RegApprox1}.

\begin{lemma}
  \label{lem:a_RegApproxTrunc}
  Suppose Assumptions~\ref{assu:EDR}-\ref{assu:source} hold and $s \leq \alpha_0$.
  Let $\lambda \asymp n^{-\theta}$ with $\theta \in (0,\beta)$ and $\delta \in (0,1)$.
  For any $t > 0$, denote $\Omega_t = \{x \in \caX: \abs{f_{\rho}^*(x)} \leq t \}$ and $\bar{\xi}(x) = \xi(x) \bm{1}_{\{x \in \Omega_{t}\}}$.
  Then, for $\alpha > \alpha_0 = \beta^{-1}$ being sufficiently close, it holds with probability at least $1-\delta$ that
  \begin{equation}
    \norm{\frac{1}{n} \sum_{i=1}^{n} \bar{\xi}(x_i) - \mathbb{E} \bar{\xi}(x)  }
    \leq C \ln  \frac{2}{\delta} \cdot \left[  \frac{M_\alpha}{n}\left( M_\alpha \lambda^{-\alpha} + t \lambda^{-\frac{\alpha+s}{2}} \right)
    + M_{\alpha} \sqrt {\frac{\lambda^{-\alpha} \ln n}{n}} \right] \lambda^{s/2}.
  \end{equation}
  Consequently, if $t \asymp n^{l}$ with $l < 1 - \frac{\alpha+s}{2}\theta$, we have
  \begin{align}
    \label{eq:a_RegApproxTrunc}
    \norm{\frac{1}{n} \sum_{i=1}^{n} \bar{\xi}(x_i) - \mathbb{E} \bar{\xi}(x) } = o_{\bbP}(\lambda^{s/2}).
  \end{align}

\end{lemma}
\begin{proof}
  We follow the same routine of the proof of \cref{lem:a_RegApprox1} and obtain \cref{eq:a_ProofApprox1_Moment} with
  $\xi$ replaced with $\bar{\xi}$.
  The only difference is that we have to control
  \begin{align*}
    \norm{\bm{1}\{x \in \Omega_{t}\} (f_{\lambda} - f_{\rho}^{*})}_{L^\infty}
    & \leq \norm{f_{\lambda}}_{L^\infty}  + \norm{\bm{1}\{x \in \Omega_{t}\}f_{\rho}^{*} }_{L^\infty} \\
    & \leq M_\alpha \norm{f_{\lambda}}_{[\caH]^\alpha} + t \\
    & \leq C M_{\alpha} \lambda^{(s-\alpha)/2} + t,
  \end{align*}
  where we apply \cref{lem:RegF_Norms} at the second inequality.
  Then, \cref{eq:a_ProofApprox1_Bernstein} changes to
  \begin{align*}
    \frac{1}{2} m! \left( C M_\alpha^2 \lambda^{\tilde{s}-\alpha} \ln n \right) \cdot \left( C M_{\alpha}^2 \lambda^{-\alpha+\tilde{s}/2} + M_\alpha \lambda^{-\alpha/2} t \right)^{m-2}
  \end{align*}
  and the rest follows.
\end{proof}

To bound the extra error terms caused by truncation, we have to use the following proposition about the $L^q$ embedding of the RKHS
~\citep[Theorem 5]{zhang2023_OptimalityMisspecified}.
\begin{proposition}
  \label{prop:LqEMB}
  Under \cref{assu:EMB}, for any $0 < s \leq \alpha_0$ and $\alpha > \alpha_0$, we have embedding
  \begin{align}
  [\caH]
    ^s \hookrightarrow L^{q_s}(\caX,\dd \mu),\quad q_s = \frac{2\alpha}{\alpha - s}.
  \end{align}
\end{proposition}

\begin{lemma}
  \label{lem:a_RegApprox2}
  Suppose Assumptions~\ref{assu:EDR}-\ref{assu:source} hold and $s \leq \alpha_0$.
  Let $\lambda \asymp n^{-\theta}$ with $\theta \in (0,\beta)$ and $\delta \in (0,1)$.
  Then
  \begin{align}
    \label{eq:a_RegApprox2}
    \norm{T_\lambda^{-\frac{1}{2}}\left[\left(\tilde{g}_Z-T_X f_\lambda\right)-\left(g-T f_\lambda\right)\right]}
    = o_{\bbP}(\lambda^{s/2}) = o_{\bbP}(n^{-s\theta/2}).
  \end{align}
\end{lemma}
\begin{proof}
  We will choose $t = n^l$ for some $l$ that will be determined later and choose some $\alpha > \alpha_0$ being sufficiently close.
  Using the same notations as in \cref{eq:a_RegApproxTrunc}, we decompose
  \begin{align}
    \label{eq:a_Proof_TruncDecomp}
    \begin{aligned}
      \norm{\frac{1}{n} \sum_{i=1}^n \xi(x_i)-\mathbb{E} \xi(x)}_{\caH}
      & \leq \norm{\frac{1}{n} \sum_{i=1}^n \bar{\xi}(x_i) -  \E \bar{\xi}(x)}_{\caH}
      + \norm{\frac{1}{n} \sum_{i=1}^n \xi(x_i) \bm{1}_{\{x_i \notin \Omega_{t}\}}}_{\caH} \\
      & \quad + \norm{\E \xi(x)\bm{1}_{\{x \notin \Omega_{t}\}}}_{\caH}.
    \end{aligned}
  \end{align}

  The first term in \cref{eq:a_RegApproxTrunc} is already bounded by \cref{eq:a_RegApproxTrunc} if $l < 1-\frac{\alpha + s}{2}\theta$.
  To bound the second term in \cref{eq:a_Proof_TruncDecomp}, we notice that
  \begin{align*}
    x_i \in \Omega_t,~\forall i = 1,\dots,n\qq{implies} \frac{1}{n} \sum_{i=1}^n \xi(x_i) \bm{1}_{\{x_i \notin \Omega_{t}\}} = 0.
  \end{align*}
  Since Markov's inequality yields
  \begin{align}
    \label{eq:a_Proof_OmegaProbControl}
    \bbP_{x\sim \mu}\left\{ x \notin \Omega_t \right\} \leq t^{-q} \norm{f^*_\rho}_{L^q}^q,
  \end{align}
  where $q = \frac{2\alpha}{\alpha - s}$,
  we get
  \begin{align*}
    \bbP\left\{ x_i \in \Omega_t,~\forall i  \right\}
    = \left( \bbP_{x\sim \mu}\left\{ x \in \Omega_t \right\} \right)^n
    = (1 - \bbP_{x\sim \mu}\left\{ x \notin \Omega_t \right\} )^n
    \geq (1 - t^{-q} \norm{f^*_\rho}_{L^q}^q )^n.
  \end{align*}
  So the second term vanishes with high probability as long as $l > 1/q$.

  For the third term in \cref{eq:a_Proof_TruncDecomp}, using \cref{eq:a_RegK_H}, we get
  \begin{align*}
    \norm{\E \xi(x)\bm{1}_{\{x \notin \Omega_{t}\}}}_{\caH}
    & \leq \E \norm{\xi(x)\bm{1}_{\{x \notin \Omega_{t}\}} }_{\caH} \\
    & = \E \left[ \bm{1}_{\{x \notin \Omega_{t}\}}(f^*_\rho(x) - f_\lambda(x)) \norm{(T_\lambda)^{-1/2}k(x,\cdot)}_{\caH} \right] \\
    & \leq M_\alpha \lambda^{-\alpha/2} \E \left[ \bm{1}_{\{x \notin \Omega_{t}\}}(f^*_\rho(x) - f_\lambda(x)) \right] \\
    & \leq M_\alpha \lambda^{-\alpha/2}  \left[\E (f^*_\rho(x) - f_\lambda(x))^2 \right]^{\frac{1}{2}} \left[\bbP\{x \notin \Omega_{t}\}\right]^{\frac{1}{2}} \\
    & \leq M_\alpha \lambda^{-\alpha/2} \lambda^{s/2} t^{-q/2} \norm{f^*_\rho}_{L^q}^{q/2}.
  \end{align*}
  Consequently, if $l > \frac{\alpha\theta}{q}$, then
  \begin{align*}
    \norm{\E \xi(x)\bm{1}_{\{x \notin \Omega_{t}\}}}_{\caH} = o(\lambda^{s/2}).
  \end{align*}

  Finally, the three requirements of $l$ are
  \begin{align*}
    l < 1 - \frac{\alpha+s}{2}\theta,\quad l > \frac{1}{q},\qq{and} l > \frac{\theta \alpha}{q},
  \end{align*}
  where $q = \frac{2\alpha}{\alpha-s}$.
  Since $\theta < \beta = \alpha_0^{-1}$, we can choose $\alpha$ sufficiently close to $\alpha_0$ such that $\theta\alpha < 1$.
  Then,
  \begin{align*}
  (1 - \frac{\alpha+s}{2}\theta)
    - \frac{1}{q} = \left( 1 - \theta\alpha  \right) \left( \frac{\alpha+s}{2\alpha} \right)  > 0,
  \end{align*}
  and thus
  \begin{align*}
    \frac{\theta \alpha}{q} < \frac{1}{q} < 1 - \frac{\alpha+s}{2}\theta,
  \end{align*}
  showing that we can choose some $l$ satisfying all the requirements and the proof is finish.
\end{proof}

\subsection{The noiseless case}

The case when $\lambda = n^{-\theta}$ for $\theta < \beta$ is already covered in \cref{thm:BiasAsymp}.
For the case $\theta \geq \beta$, the approximation \cref{lem:ConcenIneq} no longer holds, and we must reduce it to the former case.
However, there is no direct monotone property of $\mathbf{Bias}(\lambda)$.
Nevertheless, we have the following monotone relation about the operator norms,
whose proof utilizes the idea in \citet[Proposition 6.1]{lin2021_KernelInterpolation} with modification.
\begin{proposition}
  \label{prop:a_RemainingMonotone}
  Let $\psi_\lambda = \lambda (T_X +\lambda)^{-1} \in \mathscr{B}(\caH)$.
  Suppose $\lambda_1 \leq \lambda_2$, then for any $s,p \geq 0$,
  \begin{align}
    \label{eq:a_RemainingMonotone}
    \norm{T^s\psi_{\lambda_1}^p}_{\mathscr{B}(\caH)} = \norm{\psi_{\lambda_1}^p T^s}_{\mathscr{B}(\caH)}
    \leq \norm{T^s\psi_{\lambda_2}^p}_{\mathscr{B}(\caH)} = \norm{\psi_{\lambda_2}^p T^s}_{\mathscr{B}(\caH)}.
  \end{align}
\end{proposition}
\begin{proof}
  Let us denote by $\preceq$ the partial order induced by positive operators.
  Since the function $\lambda \mapsto \frac{\lambda}{z+\lambda}$ is monotone decreasing with $\lambda$, we obtain
  $\psi_{\lambda_1}^{2p} \preceq \psi_{\lambda_2}^{2p}$, which further implies
  \begin{align*}
    T^s \psi_{\lambda_1}^{2p} T^s \preceq T^s  \psi_{\lambda_2}^{2p} T^s.
  \end{align*}
  Then, since $\norm{A}^2 = \norm{A A^*}$, we have
  \begin{align*}
    \norm{T^{s}\psi_{\lambda_1}^p}_{\mathscr{B}(\caH)}^2 = \norm{ T^s \psi_{\lambda_1}^{2p} T^s}_{\mathscr{B}(\caH)}
    \leq \norm{ T^s \psi_{\lambda_2}^{2p} T^s}_{\mathscr{B}(\caH)} = \norm{T^{s}\psi_{\lambda_2}^p}_{\mathscr{B}(\caH)}^2,
  \end{align*}
  and the equality in \cref{eq:a_RemainingMonotone} is proven by $\norm{A} = \norm{A^*}$.
\end{proof}

\begin{lemma}
  \label{lem:BiasUpperInterpolating}
  Under \cref{assu:EDR},\ref{assu:EMB},\ref{assu:source}, assume further $s > 1$.
  Suppose $\lambda \asymp n^{-\theta}$ for $\theta \geq \beta$.
  Then,
  \begin{align}
    \label{eq:BiasUpperInterpolating}
    \mathbf{Bias}^2(\lambda) = O^{\poly}_{\bbP}(n^{-\min(s,2)\beta}).
  \end{align}
\end{lemma}
\begin{proof}
  Since $f^*_\rho$ is given in \cref{eq:fHs} and $s > 1$,
  we have $f^*_\rho \in [\caH]^t$ for $1 \leq t < s$.
  In particular, $f^*_\rho \in \caH$, so the bias term can also be written as
  \begin{align}
    \label{eq:BiasExplicit}
    \mathbf{Bias}(\lambda) =  \norm{\lambda (T_X+\lambda)^{-1}f^{*}_{\rho}}_{L^2}.
  \end{align}
  Moreover, from the construction \cref{eq:InterpolationSpaceDef} of $[\caH]^t$, we may assume
  $f^*_\rho = T^{t/2} g$ for some $g \in L^2$ with $\norm{g}_{L^2} \leq C$, and restrict further that $t \leq 2$.
  Let $\tilde{\lambda} \asymp n^{-l}$ for $l \in (0,\beta)$.
  Then, using the same notation in \cref{prop:a_RemainingMonotone}, we have
  \begin{align*}
    \mathbf{Bias}(\lambda) &= \norm{\psi_\lambda f^{*}_{\rho}}_{L^2} = \norm{T^{1/2} \psi_\lambda T^{\frac{t-1}{2}} \cdot T^{1/2}g}_{\caH} \\
    & \leq \norm{T^{1/2}\psi_\lambda T^{(t-1)/2}} \cdot  \norm{T^{1/2}g}_{\caH} \\
    & \leq C \norm{T^{1/2}\psi_\lambda^{1/2}} \cdot  \norm{\psi_\lambda^{1/2} T^{\frac{t-1}{2}} } \\
    & \leq C \norm{T^{1/2}\psi_{\tilde{\lambda}}^{1/2}}\cdot  \norm{\psi_{\tilde{\lambda}}^{1/2} T^{\frac{t-1}{2}} } \\
    & \leq C \norm{T^{1/2}\psi_{\tilde{\lambda}}^{1/2}} \cdot  \norm{\psi_{\tilde{\lambda}}^{(2-t)/2}}
    \cdot \norm{\psi_{\tilde{\lambda}}^{\frac{t-1}{2}}T^{\frac{t-1}{2}}} \\
    &= C \tilde{\lambda}^{t/2} \norm{\psi_{\tilde{\lambda}}^{(2-t)/2}} \cdot \norm{T^{1/2} T_{X\tilde{\lambda}}^{-1/2} }
    \cdot \norm{T^{\frac{t-1}{2}} T_{X\tilde{\lambda}}^{-\frac{t-1}{2}} } \\
    & \leq  C \tilde{\lambda}^{t/2} \norm{T^{1/2} T_{X\tilde{\lambda}}^{-1/2}}^t,
  \end{align*}
  where we use \cref{lem:OpIneq_Cordes} for the last inequality.
  Finally, since $\tilde{\lambda} \asymp n^{-l}$ for $l < \beta$, \cref{lem:ConcenIneq} implies that with high probability
  we have
  \begin{align*}
    \norm{T^{\frac12} T_{X\tilde{\lambda}}^{-\frac12}} =
    \norm{T^{\frac12}T_\lambda^{-\frac12} T_\lambda^{\frac12}  T_{X\tilde{\lambda}}^{-\frac12}}
    \leq \norm{T^{\frac12}T_\lambda^{-\frac12}}  \norm{T_\lambda^{\frac12}  T_{X\tilde{\lambda}}^{-\frac12}}  \leq 1 \cdot  \sqrt{3} = \sqrt {3}.
  \end{align*}
  Therefore, we obtain
  \begin{align*}
    \mathbf{Bias}(\lambda) = O_{\bbP}(\tilde{\lambda}^{t/2}) = O_{\bbP}(n^{-tl/2}).
  \end{align*}
  Since $t < \min(s,2)$ and $l < \beta$ can arbitrarily close, we conclude \cref{eq:BiasUpperInterpolating}.
\end{proof}

\begin{proof}[Proof of \cref{prop:BiasLower}]
  Let us denote $\caF = \left\{ f : \norm{f^*_\rho}_{[\caH]^s} \leq R \right\}$ for convenience.
  Since $f^*_\rho \in \caH$, the bias term can be given by
  \begin{align*}
    \mathbf{Bias}(\lambda) =  \norm{f^{*}_{\rho} - T_X (T_X+\lambda)^{-1} f^{*}_{\rho}}_{L^2}
    = \norm{(I - L_X)f^{*}_{\rho}}_{L^2}
  \end{align*}
  for a linear operator $L_X = T_X (T_X+\lambda)^{-1} $ on $\caH$.
  Then,
  \begin{align*}
    \sup_{f^{*}_{\rho} \in \caF} \mathbf{Bias}(\lambda)
    &= \sup_{f^{*}_{\rho}\in \caF}\norm{(I - L_X)f^{*}_{\rho}}_{L^2}
    \stackrel{(a)}{=}  \sup_{\norm{g}_{\caH} \leq R} \norm{T^{\frac{1}{2}}(I - L_X)T^{\frac{s-1}{2}} g}_{\caH} \\
    &= \sup_{\norm{g}_{\caH} \leq R} \norm{(T^{\frac{s}{2}} - T^{\frac{s}{2}} L_X T^{\frac{s-1}{2}}) g}_{\caH}  \\
    &= \norm{T^{\frac{s}{2}} - T^{\frac{s}{2}} L_X T^{\frac{s-1}{2}}}_{\mathscr{B}(\caH)} \\
    & \stackrel{(b)}{\geq} \lambda_{n+1}^{s/2} = (n+1)^{-s\beta/2} = \Omega(n^{-s\beta/2}),
  \end{align*}
  where in (a) we use the relation between the interpolation spaces and in (b) we use the fact that
  and $\norm{A-B} \geq \lambda_{n+1}(A)$ for any operator $B$ with rank at most $n$
  (see, for example, \citet[Section 3.5]{simon2015_OperatorTheorya}).

\end{proof}

  \section{Auxiliary results}\label{sec:auxiliary-results}
\begin{proposition}
  \label{prop:a_KRR_Filter}
  Let
  \begin{align*}
    f(z) = \frac{z^\alpha}{z + \lambda}.
  \end{align*}
  Then,
  \begin{enumerate}[(1)]
    \item If $\alpha = 0$, then $f(z)$ is monotone decreasing.
    \item If $\alpha \in (0,1)$, then $f(z)$ is monotone increasing in $[0, \frac{\alpha \lambda}{1-\alpha}]$,
    and decreasing in $[\frac{\alpha \lambda }{1-\alpha},+\infty)$.
    Consequently, $f(z) \leq \lambda^{\alpha-1}$.
    \item If $\alpha \geq 1$, then $f(z)$ monotone increasing on $[0,+\infty)$.
  \end{enumerate}
\end{proposition}
\begin{proof}
  We simply notice that
  \begin{align*}
    f'(z) = \frac{z^{\alpha-1}}{(z+\lambda)^2}(\alpha \lambda - (1-\alpha)z).
  \end{align*}
\end{proof}

\begin{proposition}
  \label{prop:SeriesAsymp}
  Suppose  $c_\beta i^{-\beta}\leq \lambda_i \leq C_\beta i^{-\beta}$ and $p > 0$, then as $\lambda \to 0$, we have
  \begin{align}
    \sum_{i = 1}^\infty  \left( \frac{\lambda_i^{p} }{\lambda_i + \lambda} \right)^2 i^{-1}
    \asymp
    \begin{cases}
      \lambda^{2(p-1)}, & p < 1; \\
      \ln \frac{1}{\lambda}, & p = 1; \\
      1, & p > 1.
    \end{cases}
  \end{align}
\end{proposition}
\begin{proof}
  We first consider the case when $p < 1$.
  Since $c_\beta i^{-\beta}\leq \lambda_i \leq C_\beta i^{-\beta}$, from \cref{prop:a_KRR_Filter}, letting $q = \frac{p}{1-p}$,
  we have
  \begin{align*}
    \frac{\lambda_i^{p} }{\lambda_i + \lambda} \leq
    \begin{cases}
      \frac{C_\beta^{p} i^{-p\beta} }{C_\beta i^{-\beta} + \lambda}, & \qif C_\beta i^{-\beta} \leq q \lambda; \\
      \lambda_i^{p} / \lambda_i \leq C_\beta^{p} i^{-(p-1)\beta} , & \qif C_\beta i^{-\beta} > q \lambda; \\
    \end{cases}
  \end{align*}
  Therefore,
  \begin{align*}
    \sum_{i = 1}^\infty  \left( \frac{ \lambda_i^{p} }{\lambda_i + \lambda} \right)^2 i^{-1}
    &\leq C \sum_{i: C_\beta i^{-\beta} > q \lambda}  i^{-2(p-1)\beta-1} +
    C \sum_{i: C_\beta i^{-\beta} \leq q \lambda}\frac{i^{-2p \beta} }{(C_\beta i^{-\beta} + \lambda)^2} i^{-1} \\
    & \eqqcolon S_1 + S_2.
  \end{align*}
  For $S_1$, noticing $C_\beta i^{-\beta} > q \lambda$ implies $i < (q\lambda/C_\beta)^{-1/\beta}$, we have
  \begin{align*}
    S_1 \leq C \sum_{i=1}^{\lfloor (q\lambda/C_\beta)^{-1/\beta}\rfloor}i^{-2(p-1)\beta-1} \leq C\lambda^{2(p-1)}.
  \end{align*}
  For $S_2$, using \cref{prop:a_KRR_Filter} again we have
  \begin{align*}
    S_2 &\leq C \int_{(q\lambda/C_\beta)^{-1/\beta}-1}^\infty \frac{x^{-2p\beta} }{(C_\beta x^{-\beta} + \lambda)^2} x^{-1} \dd x \\
    & = C \lambda^{2p-2} \int_{(C_\beta/q)^{1/\beta} -\lambda^{1/\beta}}^{\infty} \frac{ y^{-2p\beta}}{(C_\beta y^{-\beta} + 1)^2}y^{-1}\dd y
    \quad (x = \lambda^{-1/\beta} y)  \\
    & \leq C \lambda^{2p-2},
  \end{align*}
  where we note that the last integral is bounded above by a constant.
  Therefore, we conclude that $\norm{f_\lambda - f_{\rho}^*}_{[\mathcal{H}]^{\gamma}}^2 \leq C \lambda^{2p-2}$.
  For the lower bound, if $C_\beta i^{-\beta} \leq q\lambda$, we have
  \begin{align*}
    \frac{\lambda_i^{p}}{\lambda_i + \lambda} \geq \frac{c_\beta^{p} i^{-p\beta} }{c_\beta i^{-\beta} + \lambda},
  \end{align*}
  and hence
  \begin{align*}
    \sum_{i = 1}^\infty  \left( \frac{\lambda_i^{p} }{\lambda_i + \lambda} \right)^2 i^{-1}
    & \geq C  \int_{(q\lambda/C_\beta)^{-1/\beta}}^\infty \frac{x^{-2p\beta} }{(C_\beta x^{-\beta} + \lambda)^2} x^{-1} \dd x \\
    & = C\lambda^{2p-2} \int_{(C_\beta/p)^{1/\beta}}^{\infty} \frac{ y^{-2p\beta}}{(C_\beta y^{-\beta} + 1)^2}y^{-1}\dd y \\
    & \geq C \lambda^{2p-2},
  \end{align*}
  where we note that the last integral is independent of $\lambda$.

  For the case $p = 1$, by \cref{prop:a_KRR_Filter}, we have
  \begin{align*}
    \sum_{i = 1}^\infty  \left( \frac{\lambda_i}{\lambda_i + \lambda} \right)^2 i^{-1}
    &\leq C \sum_{i = 1}^\infty  \left( \frac{i^{-\beta}}{C_\beta i^{-\beta} + \lambda} \right)^2 i^{-1} \\
    & \leq C \sum_{i = 1}^{\lfloor 2 \lambda^{-1/\beta}\rfloor} \left( \frac{i^{-\beta}}{C_\beta  i^{-\beta} + \lambda} \right)^2 i^{-1}
    + C \sum_{i = \lfloor 2 \lambda^{-1/\beta}\rfloor+1}^{\infty } \left( \frac{i^{-\beta}}{C_\beta  i^{-\beta} + \lambda} \right)^2 i^{-1} \\
    & \leq C \sum_{i = 1}^{\lfloor 2 \lambda^{-1/\beta}\rfloor} i^{-1} +
    C\int_{2 \lambda^{-1/\beta}}^{\infty} \left( \frac{x^{-\beta}}{C_\beta x^{-\beta} + \lambda} \right)^2 x^{-1} \dd x \\
    & \leq C \ln \frac{1}{\lambda}  + C \int_{2}^\infty \left( \frac{y^{-\beta}}{C_\beta y^{-\beta} + 1} \right)^2 y^{-1} \dd y\\
    & \leq C \ln \frac{1}{\lambda}.
  \end{align*}
  For the lower bound, we have
  \begin{align*}
    \sum_{i = 1}^\infty  \left( \frac{\lambda_i}{\lambda_i + \lambda} \right)^2 i^{-1}
    & \geq c  \sum_{i = 1}^{\lfloor \lambda^{-1/\beta}\rfloor} \left( \frac{i^{-\beta}}{c_\beta i^{-\beta} + \lambda} \right)^2 i^{-1} \\
    & \geq c \sum_{i = 1}^{\lfloor \lambda^{-1/\beta}\rfloor} i^{-1}
    \geq c \ln \frac{1}{\lambda}.
  \end{align*}

  For the case $p > 1$, by \cref{prop:a_KRR_Filter}, we have
  \begin{align*}
    \sum_{i = 1}^\infty  \left( \frac{\lambda_i^{p} }{\lambda_i + \lambda} \right)^2 i^{-1}
    \leq C  \sum_{i = 1}^\infty \frac{i^{-2p\beta} }{(C_\beta i^{-\beta} + \lambda)^2} i^{-1}
    & \leq C \sum_{i = 1}^\infty  i^{-2(p-1)\beta - 1 }
    \leq C,
  \end{align*}
  since the last series is summable.
  The lower bound is derived by
  \begin{align*}
    \sum_{i = 1}^\infty  \left( \frac{\lambda_i^{p} }{\lambda_i + \lambda} \right)^2 i^{-1}
    \geq \frac{\lambda_1^{p} }{\lambda_1 + \lambda} \geq c.
  \end{align*}
\end{proof}

\begin{proposition}
  \label{prop:EffectiveDimEstimation}
  Under \cref{assu:EDR}, for any $p \geq 1$, we have
  \begin{align}
    \label{eq:EffectDim}
    \mathcal{N}_p(\lambda)= \tr \left( TT_\lambda^{-1} \right)^p =
    \sum_{i = 1}^\infty \left( \frac{\lambda_i}{\lambda + \lambda_i} \right)^p \asymp \lambda^{-1/\beta}.
  \end{align}
\end{proposition}
\begin{proof}
  Since $c ~i^{-\beta} \leq \lambda_i \leq C i^{-\beta}$, we have
  \begin{align*}
    \mathcal{N}_{p}(\lambda) &= \sum_{i = 1}^{\infty} \left( \frac{\lambda_i}{\lambda_i + \lambda}  \right)^p
    \leq \sum_{i = 1}^{\infty} \left( \frac{C i^{-\beta}}{C i^{-\beta} + \lambda} \right)^p = \sum_{i = 1}^{\infty} \left( \frac{C }{C+ \lambda i^{\beta}}  \right)^p \\
    &\leq \int_{0}^{\infty} \left( \frac{C}{\lambda x^{\beta} + C} \right)^p\dd x
    = \lambda^{-1/\beta} \int_{0}^{\infty}  \left(\frac{C }{y^{\beta} + C} \right)^p\dd y \leq \tilde{C} \lambda^{-1/\beta}.
  \end{align*}
  for some constant $C$.
  The lower bound is similar.
\end{proof}

The following inequality about vector-valued random variables is well-known in the literature~\citep{caponnetto2007_OptimalRates}.
\begin{lemma}
  \label{lem:ConcenHilbert}
  Let $H$ be a real separable Hilbert space.
  Let $\xi,\xi_1,\dots,\xi_n$ be i.i.d.\ random variables taking values in $H$.
  Assume that
  \begin{align}
    \label{eq:HilbertConcenCondition}
    \E \norm{\xi - \E \xi}_{H}^m \leq \frac{1}{2}m!\sigma^2 L^{m-2},\quad \forall m = 2,3,\dots.
  \end{align}
  Then for fixed $\delta \in (0,1)$, one has
  \begin{align}
    \bbP \left\{ \norm{\frac{1}{n}\sum_{i=1}^n \xi_i - \E \xi}_{H}
    \leq 2\left( \frac{L}{n} + \frac{\sigma}{\sqrt{n}} \right) \ln \frac{2}{\delta} \right\}
    \geq 1- \delta.
  \end{align}
  Particularly, a sufficient condition for \cref{eq:HilbertConcenCondition} is
  \begin{align*}
    \norm{\xi}_{H} \leq \frac{L}{2}\  \text{a.s., and}\
    \E \norm{\xi}_H^2 \leq \sigma^2.
  \end{align*}
\end{lemma}

The following concentration result has been shown in \citet{fischer2020_SobolevNorm,zhang2023_OptimalityMisspecified}.
We use the form in \citet[Proposition 5.8]{li2023_KernelInterpolation} for convenience,
see also \citet[Lemma 12]{zhang2023_OptimalityMisspecified}.

\begin{lemma}
  \label{lem:ConcenIneq}
  Suppose $\caH$ has embedding index $\alpha_0$ and \cref{assu:EDR} holds.
  Let $\lambda = \lambda(n) \to 0$ satisfy
  $\lambda = \Omega\left(n^{-1/\alpha_0 + p}\right)$ for some $p > 0$
  and fix arbitrary $\alpha \in (\alpha_0,\alpha_0 + p)$.
  Then, for all $\delta \in (0,1)$,
  when $n$ is sufficiently large, with probability at least $1 - \delta$,
  \begin{equation}
    \norm{T_{\lambda}^{-\frac{1}{2}} (T - T_X) T_{\lambda}^{-\frac{1}{2}} }_{\mathscr{\caH}}
    \leq C \left( \frac{\lambda^{-\alpha}}{n} \ln n \right)^{1/2},
  \end{equation}
  where $C > 0$ is a constant no depending on $\delta, n,\alpha$,
  and we also have
  \begin{align}
    \label{eq:a_RatioControl}
    \norm{T_{\lambda}^{1/2} T_{X\lambda}^{-1/2}}_{\mathscr{B}(\caH)},~\norm{T_{\lambda}^{-1/2} T_{X\lambda}^{1/2} }_{\mathscr{B}(\caH)} \leq \sqrt{3}.
  \end{align}
\end{lemma}

The following operator inequality\citep{fujii1993_NormInequalities} will be used in our proofs.
\begin{lemma}[Cordes' Inequality]
  \label{lem:OpIneq_Cordes}
  Let $A,B$ be two positive semi-definite bounded linear operators on separable Hilbert space $H$.
  Then
  \begin{align}
    \norm{A^s B^s}_{\mathscr{B}(H)}
    \leq \norm{AB}_{\mathscr{B}(H)}^s,\quad \forall s \in [0,1].
  \end{align}
\end{lemma}

The following lemma is a consequence of the fact that $x^r$ is operator monotone when $r \in (0,1]$ and is Lipschitz when $r > 1$,
see \citet[Lemma 35]{zhang2023_OptimalityMisspecified} or \citet[Lemma 5.8]{lin2018_OptimalRates}.
\begin{lemma}
  \label{lem:OpLip}
  Suppose that $A$ and $B$ are two positive self-adjoint operators on some Hilbert space, then
  \begin{itemize}
    \item for $ r \in (0,1]$, we have
    \begin{displaymath}
      \norm{A^r-B^r} \leq \norm{A-B}^r.
    \end{displaymath}
    \item for $r \ge 1 $, denote $c=\max(\norm{A},\norm{B})$, we have
    \begin{displaymath}
      \norm{A^r-B^r} \leq r c^{r-1}\norm{A-B}.
    \end{displaymath}
  \end{itemize}
\end{lemma}


  \section{Experiments}\label{sec:details-experiments}
  \subsection{Details of experiments in the main text}
Recall that in the experiments section of the main text, we considered the kernel $k(x,y) = \min(x,y)$ and $x  \sim \mathcal{U}[0,1]$.
We know the eigensystem of $k$ that $\lambda_i = \left( \frac{2i-1}{2}\pi \right)^{-2}$ and $e_i(x) = \sqrt{2} \sin(\frac{2i-1}{2}\pi x)$.
For the three target functions used in the experiments,
simple calculation shows that the relative smoothness (source condition) of $\cos(2 \pi x), \sin(2 \pi x), \sin(\frac{3}{2}\pi x)$ are $0.5, 1.5, \infty$ respectively.

\begin{figure}[htbp]
  \centering
  \begin{minipage}{0.49\linewidth}
    \centering
    \includegraphics[width=1\linewidth]{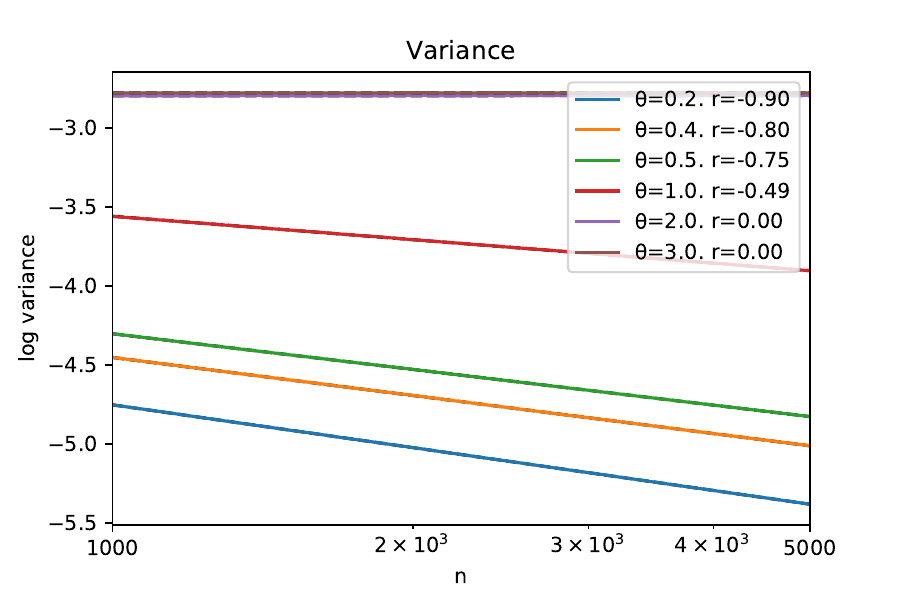}
  \end{minipage}
  \begin{minipage}{0.49\linewidth}
    \centering
    \includegraphics[width=1\linewidth]{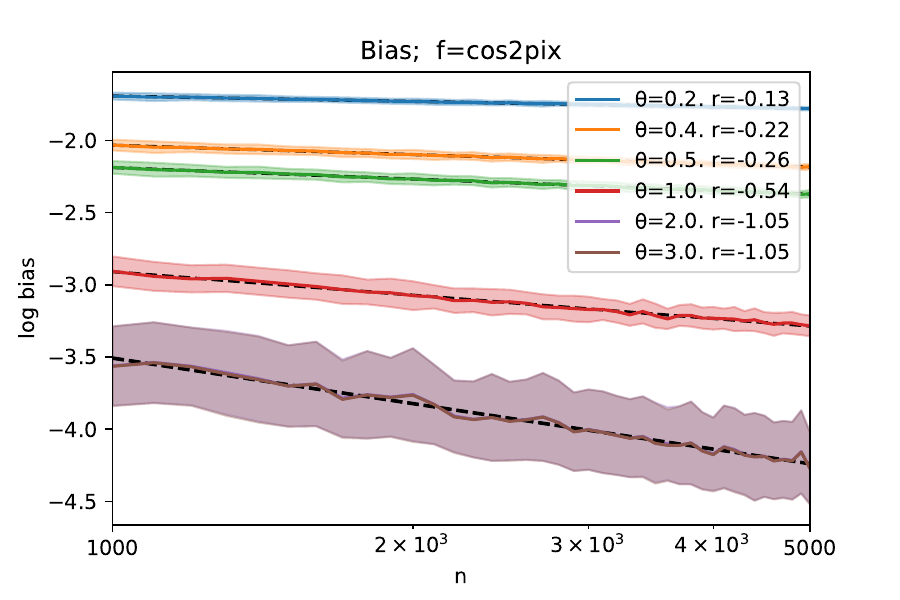}
  \end{minipage}

  \begin{minipage}{0.49\linewidth}
    \centering
    \includegraphics[width=1\linewidth]{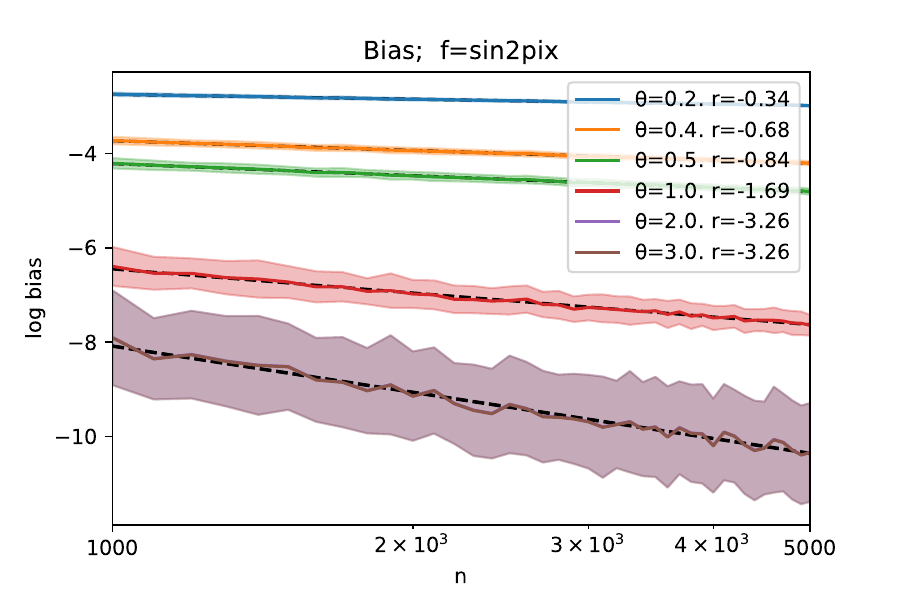}
  \end{minipage}
  \begin{minipage}{0.49\linewidth}
    \centering
    \includegraphics[width=1\linewidth]{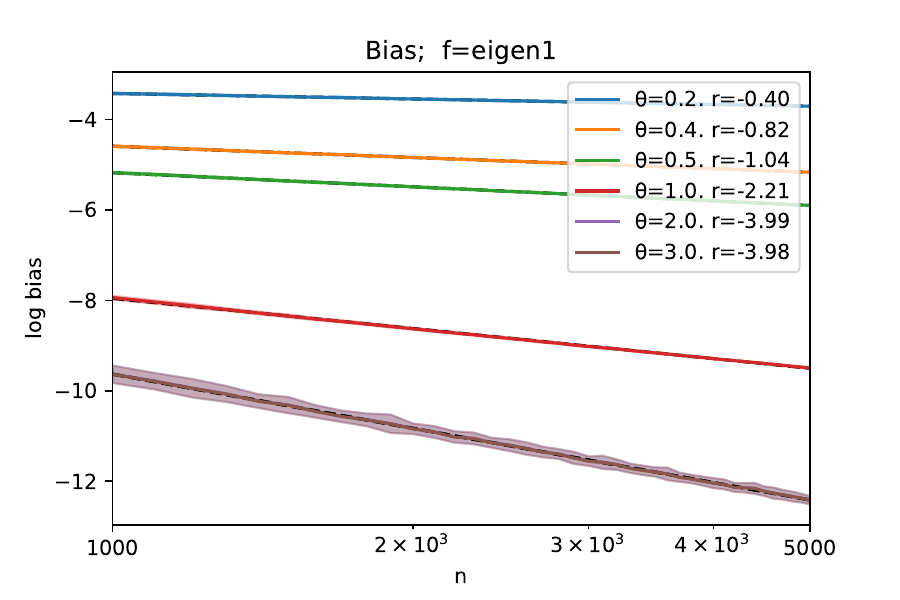}
  \end{minipage}

  \begin{minipage}{0.49\linewidth}
    \centering
    \includegraphics[width=1\linewidth]{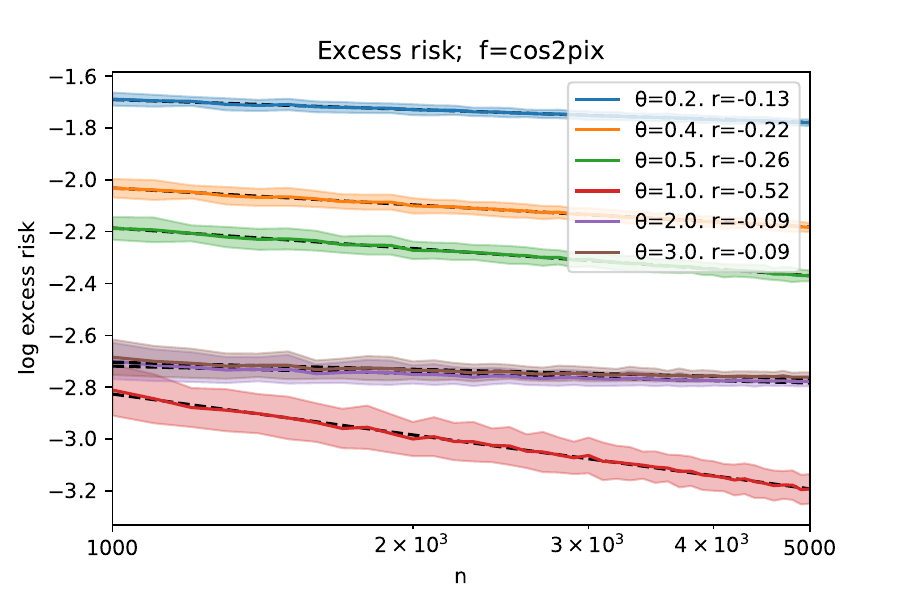}
  \end{minipage}
  \begin{minipage}{0.49\linewidth}
    \centering
    \includegraphics[width=1\linewidth]{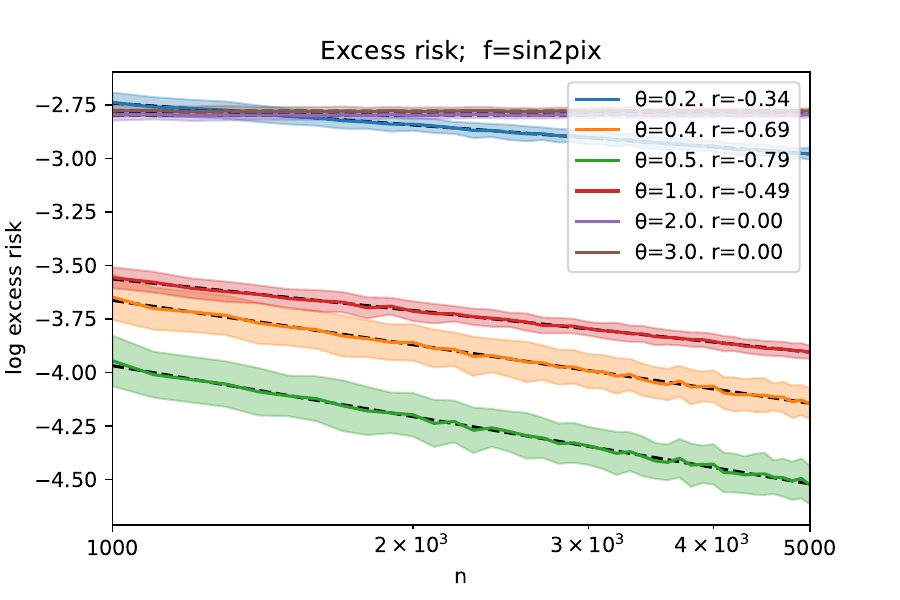}
  \end{minipage}

  \begin{minipage}{0.49\linewidth}
    \centering
    \includegraphics[width=1\linewidth]{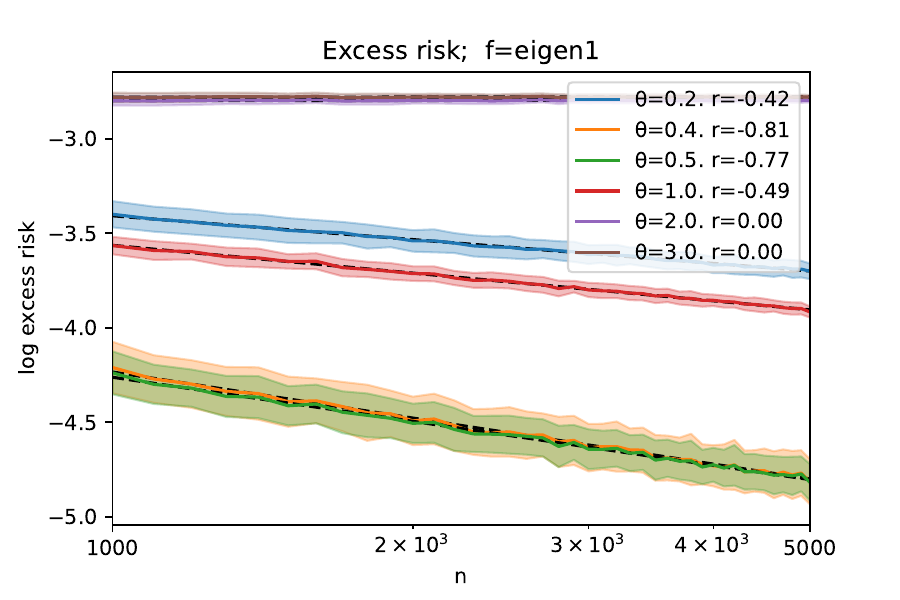}
  \end{minipage}

  \caption{Decay curves of the variance; the bias and excess risk of three target functions. Both axes are logarithmic. The curves show the average bias over 100 trials; and the regions within one standard deviation are shown in the corresponding colors.}

  \label{fig:var-bias}
\end{figure}

For some $f^*$, we generate data from the model $y = f^*(x) + \ep$ where $\ep \sim \mathcal{N}(0,0.05)$
and perform KRR with $\lambda = c n^{-\theta}$ for different $\theta$'s with some fixed constant $c$.
Then, we numerically compute the variance, bias and excess risk by Simpson's formula with $N \gg n$ nodes.
Repeating the experiment for $n$ ranged in 1000 to 5000 with an increment of 100, we can estimate the convergence rate $r$ by a logarithmic least-squares
$\log \text{err} = r \log n + b$ on the resulting values (variance, bias and excess risk).
\cref{fig:var-bias} shows the corresponding curves of the results in Table 1 in the main text.
Note that for each setting, we tried different $c$'s in the regularization parameter $\lambda = c n ^{-\theta}$ and show the curves under the best choice of $c$ ($c=0.005$).

\subsection{Learning curves with different noises}
\begin{figure}[htbp]
  \centering
  \begin{minipage}{0.49\linewidth}
    \centering
    \includegraphics[width=1\linewidth]{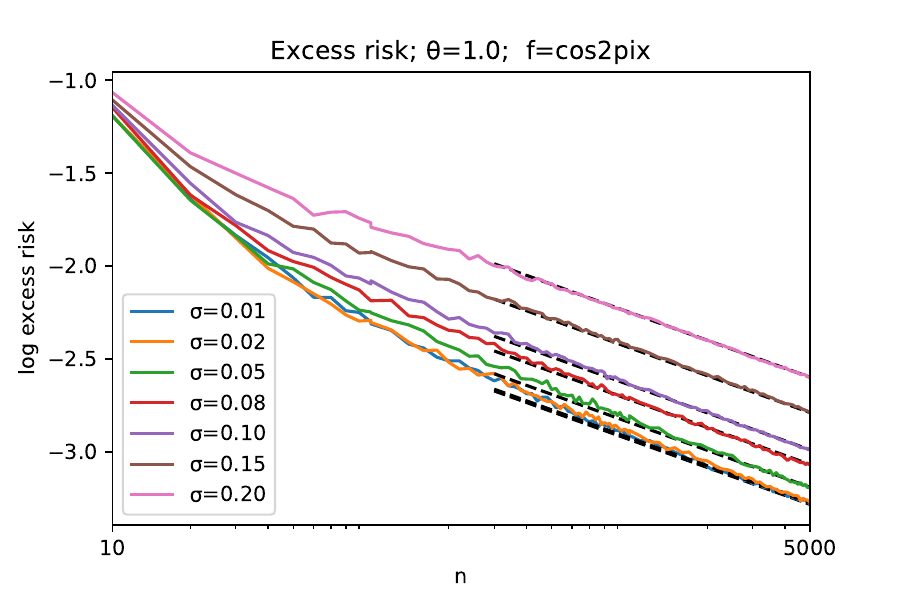}
  \end{minipage}
  \begin{minipage}{0.49\linewidth}
    \centering
    \includegraphics[width=1\linewidth]{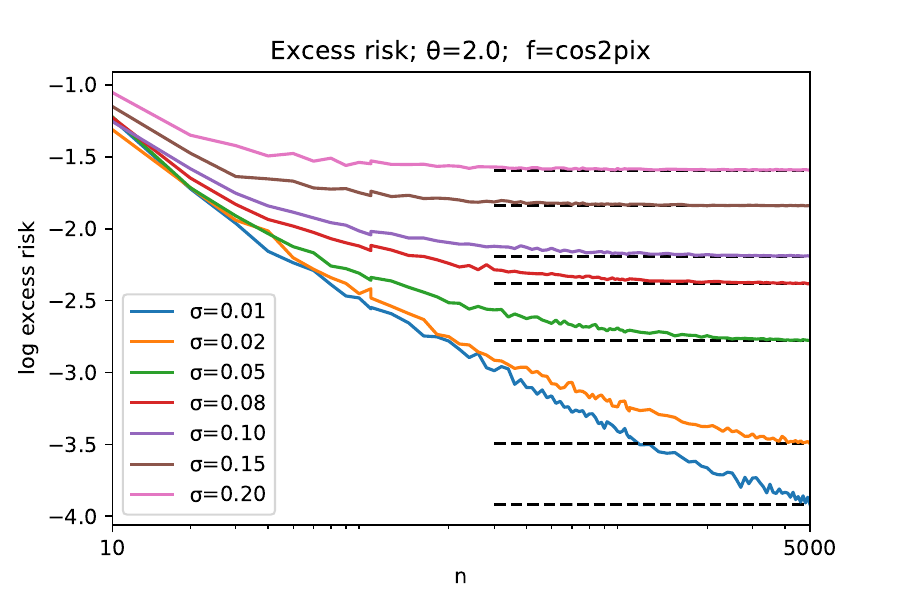}
  \end{minipage}

  \begin{minipage}{0.49\linewidth}
    \centering
    \includegraphics[width=1\linewidth]{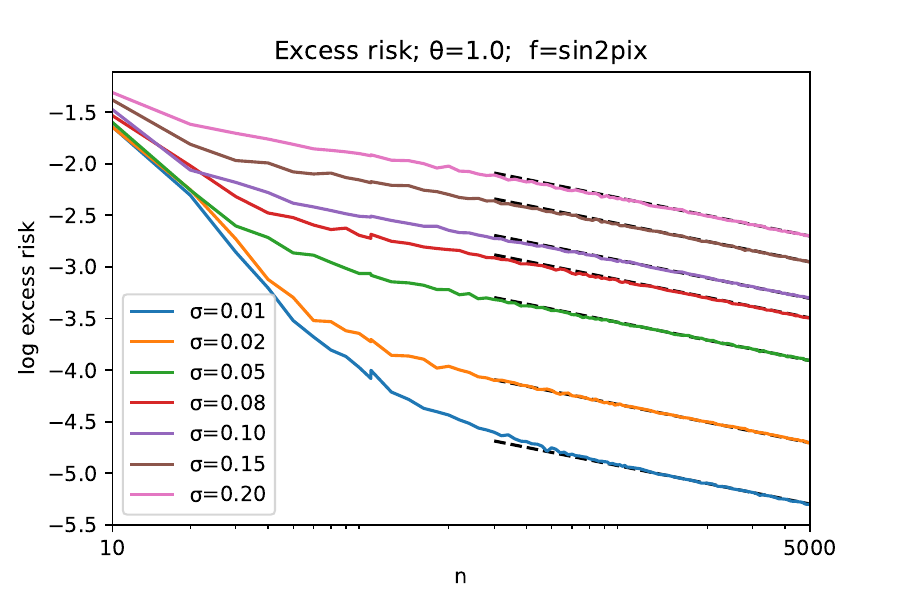}
  \end{minipage}
  \begin{minipage}{0.49\linewidth}
    \centering
    \includegraphics[width=1\linewidth]{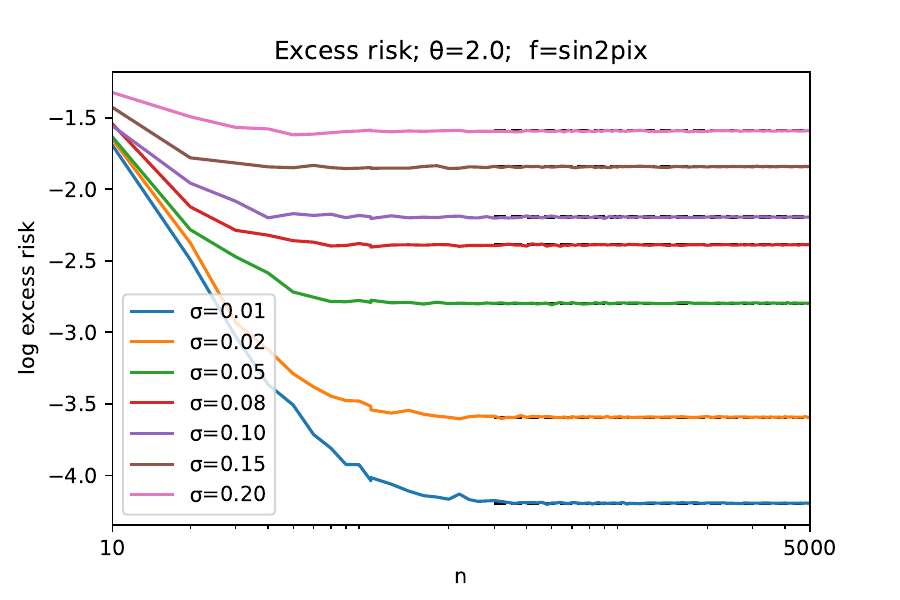}
  \end{minipage}

  \begin{minipage}{0.49\linewidth}
    \centering
    \includegraphics[width=1\linewidth]{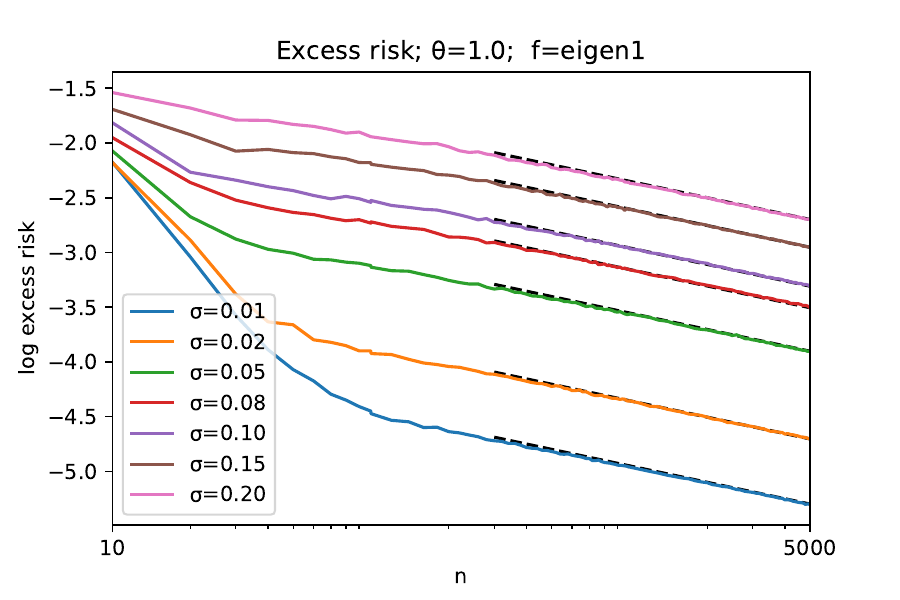}
  \end{minipage}
  \begin{minipage}{0.49\linewidth}
    \centering
    \includegraphics[width=1\linewidth]{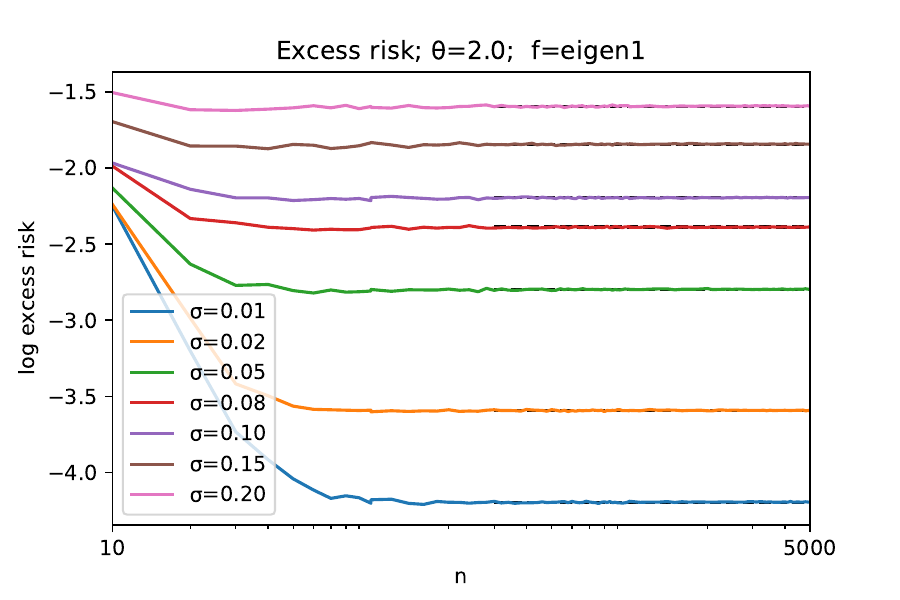}
  \end{minipage}

  \caption{Learning curves of three target functions with different noises when choosing $\lambda = c n ^{-\theta}$, $\theta = 1.0, 2.0$. Both axes are logarithmic. The black dashed lines represent the theoretical slopes under each choice of $\theta$.}

  \label{fig:noises}
\end{figure}

\cite{cui2021_GeneralizationError} discussed the `crossover from the noiseless to noisy regime' and shown the interaction between the magnitude of noise and the sample size.
As discussed in Remark 3.2 in the main text, our theory can also reflect this interaction.
In \cref{fig:noises}, we exhibit the learning curves with different magnitudes of noises and visualize this interaction.
Note that in the following the sample size is chosen as $10, 20, \cdots, 100, 120, \cdots, 1000, 1100, \cdots, 5000$, and we use the same kernel and data generation process as before.
We repeat the experiments for 100 times for each sample size and present the average excess risk.

In the above settings, the bias decays faster than variance.
\cref{fig:noises} shows that the excess risk decays fast when $n$ is relatively small and coincides the theoretical asymptotic rate in \cref{thm:LearningCurve_Noisy} when $n$ is large.
The crossover happens for smaller $n$ when the magnitude of noise is larger.
Similar phenomenon has also been reported by \citet[FIG.2, FIG.3]{cui2021_GeneralizationError}.
In addition, comparing the sample size when crossover happens for three target functions,
our results show that the crossover happens for smaller $n$ when the function is smoother, which is also consistent with \cref{thm:LearningCurve_Noisy}.

\begin{figure}[htbp]
  \centering
  \begin{minipage}{0.32\linewidth}
    \centering
    \includegraphics[width=1\linewidth]{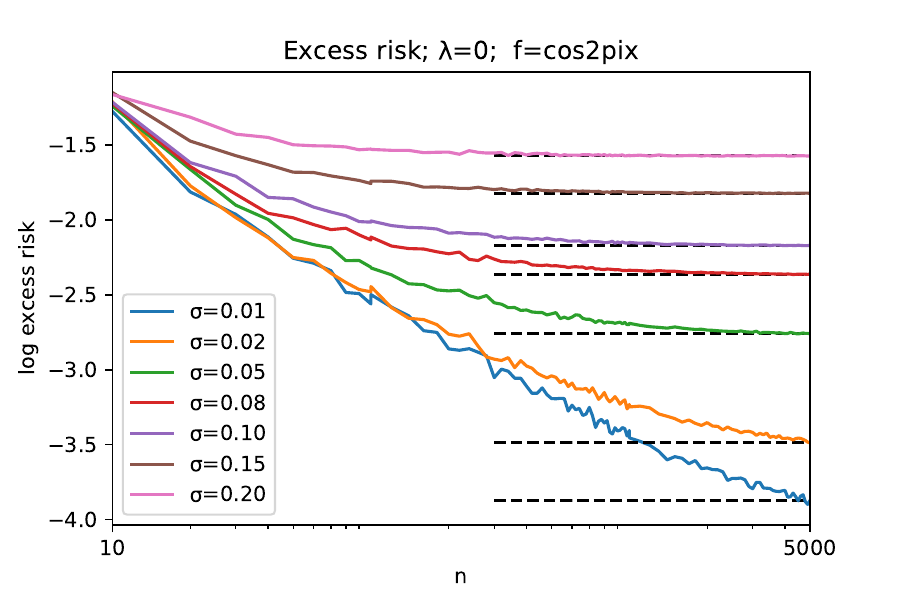}
  \end{minipage}
  \begin{minipage}{0.32\linewidth}
    \centering
    \includegraphics[width=1\linewidth]{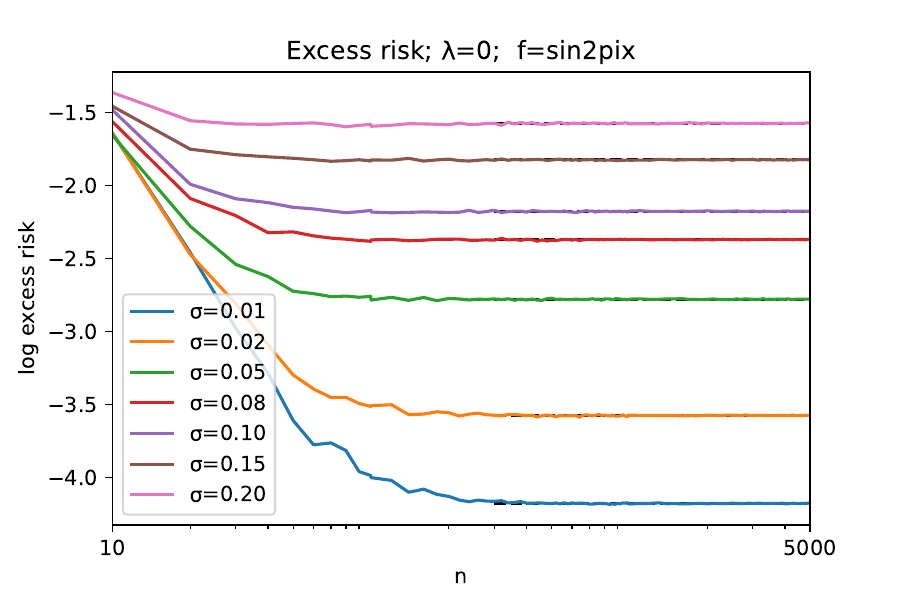}
  \end{minipage}
  \begin{minipage}{0.32\linewidth}
    \centering
    \includegraphics[width=1\linewidth]{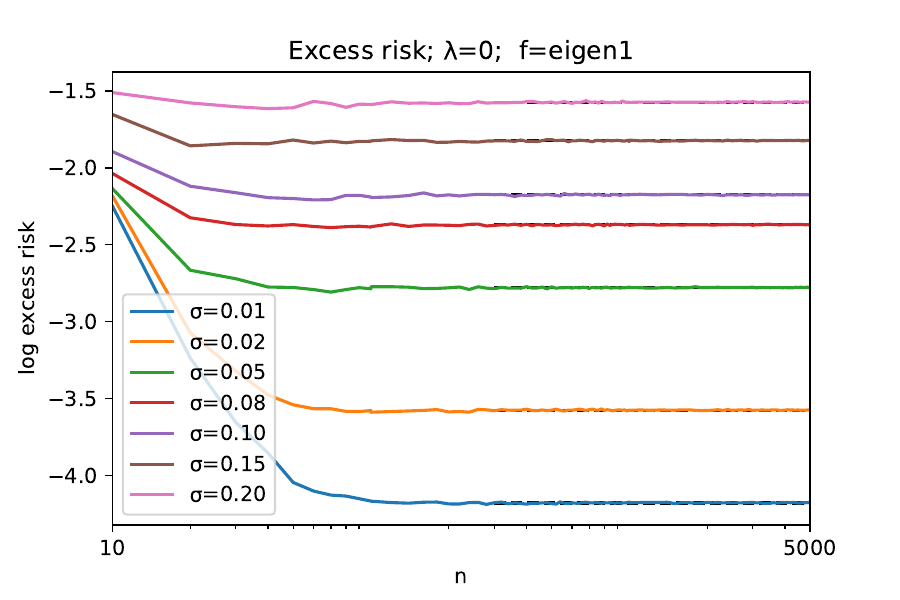}
  \end{minipage}

  \caption{Learning curves of three target functions with different noises when choosing $\lambda = 0$. Both axes are logarithmic. The black dashed lines represent the theoretical slopes.}

  \label{fig:noises_inter}
\end{figure}

\cref{thm:LearningCurve_Noisy} shows that when $\theta \ge \beta$, the excess risk is a constant asymptotically.
\cref{fig:noises_inter} shows the curves of kernel interpolation ($\lambda = 0$).
It can be seen that they are similar to the curves in the second column of \cref{fig:noises}, where we choose $ \theta = \beta = 2$.


  \bibliographystyle{plainnat}
  \bibliography{LearningCurve}



\end{document}